\def\UseGoogleArxiv{1}
\def\ArxivAuthors{%
  \setcounter{Maxaffil}{1}%
  \author{Yifan Zhang\textsuperscript{*}}%
  \author{Nghi D. Q. Bui\textsuperscript{*,\textdagger}}%
  \author{Georgios Evangelopoulos}%
  \author{Arnaud Benard}%
  \affil{Google Labs}%
  \renewcommand{\paperbrandlogo}{\includegraphics[width=105pt,trim=531bp 543bp 531bp 490bp,clip]{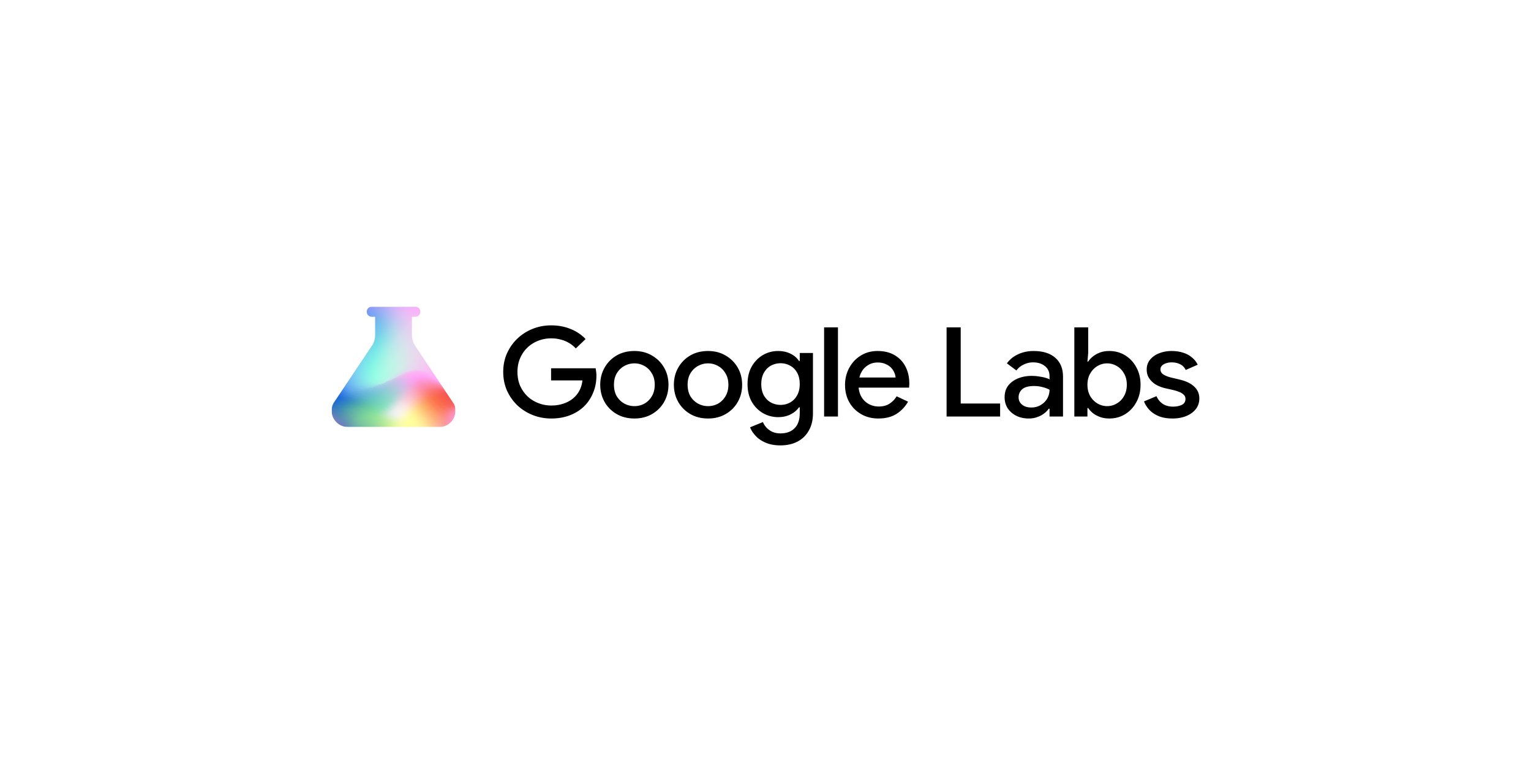}}%
  \renewcommand{\copyrightext}{\footerfont * Equal contribution.\quad \textsuperscript{\textdagger} Corresponding author: \href{mailto:nghib@google.com}{nghib@google.com}.}%
}
\newif\ifuseneurips
\newif\ifusegoogle
\newif\ifuseacl
\ifdefined\UseGoogleArxiv
  \usegoogletrue
  \ifdefined\UseACL
    \errmessage{Choose only one format: ACL or Google}
  \fi
  \ifdefined\UseNeurIPS
    \errmessage{Choose only one format: NeurIPS or Google}
  \fi
\else
  \ifdefined\UseNeurIPS
    \useneuripstrue
    \ifdefined\UseACL
      \errmessage{Choose only one format: ACL or NeurIPS}
    \fi
  \else
    \useacltrue
  \fi
\fi

\ifuseacl
  \documentclass[11pt]{article}
  \ifdefined\ACLFinal
    \usepackage[final]{acl}
  \else
    \usepackage[review]{acl}
  \fi
  \usepackage{times}
  \usepackage{latexsym}
  \usepackage[T1]{fontenc}
  \usepackage[utf8]{inputenc}
  \usepackage{inconsolata}
  \newcommand{\paperbibliographystyle}{acl_natbib}
\else
\ifuseneurips
  \documentclass{article}
  \usepackage[dblblindworkshop]{neurips_2026}
  \workshoptitle{Evaluation of Interactive Agents}
  \usepackage[utf8]{inputenc}
  \usepackage[T1]{fontenc}
  \setcitestyle{authoryear,square,comma}
  \newcommand{\paperbibliographystyle}{plainnat}
\else
  \PassOptionsToPackage{authoryear,sort&compress,round}{natbib}
  \documentclass[10pt,a4paper,logo,copyright]{googlecloud}
  \usepackage{natbib}
  \setcitestyle{authoryear,round,semicolon}
  \newcommand{\paperbibliographystyle}{abbrvnat}
\fi

\fi %

\usepackage{graphicx}
\usepackage{amsmath}
\usepackage{amssymb}
\usepackage{booktabs}
\usepackage{float}
\usepackage{tabularx}
\usepackage{enumitem}
\ifuseneurips
  \PassOptionsToPackage{expansion=false}{microtype}
\fi
\usepackage{microtype}
\usepackage{multirow}
\newcommand{\keycell}[1]{{\bfseries\boldmath #1}}
\usepackage{hyperref}
\usepackage{url}
\usepackage{xcolor}

\ifuseneurips
  \hypersetup{
    colorlinks=true,
    citecolor=blue,
    linkcolor=blue,
    urlcolor=blue
  }
\fi

\ifuseacl
  \ifdefined\ACLFinal\else
  \fi
\fi
\ifusegoogle
\fi
\ifusegoogle\else
  \title{Enabling Creative Exploration for Vibe Design Agents}
\fi
\ifusegoogle

  \title{Enabling Creative Exploration\\for Vibe Design Agents}
\fi

\ifusegoogle
\fi

\ifusegoogle\else
  \ifdefined\ACLFinal
    \author{
      \textbf{Yifan Zhang\textsuperscript{*}},
      \textbf{Nghi D. Q. Bui\textsuperscript{*,\textdagger}},
      \textbf{Georgios Evangelopoulos},
      \textbf{Arnaud Benard}
      \\
      Google
      \\
      \texttt{\{evanna, nghib, gevang, arnaudbenard\}@google.com}
      \\
      \small{
        \textsuperscript{*}Equal contribution.\quad \textsuperscript{\textdagger}\textbf{Correspondence:} \href{mailto:nghib@google.com}{nghib@google.com}
      }
    }
  \else
    \author{Anonymous Authors}
  \fi
\fi
\ifusegoogle
  \ifdefined\ArxivAuthors
    \ArxivAuthors
  \fi
\fi

\ifuseacl
  \newcommand{\paperfloatplacement}[1]{%
    \def\requestedplacement{#1}\def\hereplacement{H}%
    \ifx\requestedplacement\hereplacement
      \def\requestedplacement{t}%
    \fi}
  \newenvironment{paperfigure}[1][t]
    {\paperfloatplacement{#1}%
     \edef\startpaperfloat{\noexpand\begin{figure*}[\requestedplacement]}%
     \startpaperfloat}
    {\end{figure*}}
  \newenvironment{papertable}[1][t]
    {\paperfloatplacement{#1}%
     \edef\startpaperfloat{\noexpand\begin{table*}[\requestedplacement]}%
     \startpaperfloat}
    {\end{table*}}
\else
  \newenvironment{paperfigure}[1][t]{\begin{figure}[#1]}{\end{figure}}
  \newenvironment{papertable}[1][t]{\begin{table}[#1]}{\end{table}}
\fi

\newcommand{\paperabstract}{%
Vibe design agents turn natural-language prompts into rendered interfaces, yet repeated runs of the same request often collapse onto nearly identical palettes, fonts, and styling. Raising token temperature adds randomness directly to code generation and can hurt layout coherence. We separate visual exploration from coding by adding a lightweight planning step before implementation. Drawing on Verbalized Sampling, the agent proposes three structured design options (coordinated UI themes or image prompts) along with weights reflecting how conventional each option is. A sampler draws one option at a tunable temperature, and the downstream generator builds the UI under fixed decoding settings. Across 168 benchmark prompts, theme sampling expands repeated runs from one chosen option to nearly three and substantially increases visual variety across rendered screens. Image-prompt sampling barely changes whole-screen similarity, yet reaches a 1.32 win/loss ratio under an LLM judge, showing that localized imagery strongly shapes perceived appeal even when the rest of the layout stays similar. Finally, we evaluate joint theme and image sampling in a 54-day production A/B study (3.39 million tasks, 87,416 labeled conversations) spanning a generator upgrade, documenting steady correction rates, fewer thumbs-down ratings with wide uncertainty, and measurable latency and server-error costs.
}

\ifusegoogle
  \begin{abstract}
    \paperabstract
  \end{abstract}
\fi

\begin{document}

\maketitle

\ifusegoogle\else
  \begin{abstract}
    \paperabstract
  \end{abstract}
\fi

\section{Introduction}
\label{sec:introduction}

\begin{paperfigure}[H]
\centering
\includegraphics[width=\textwidth]{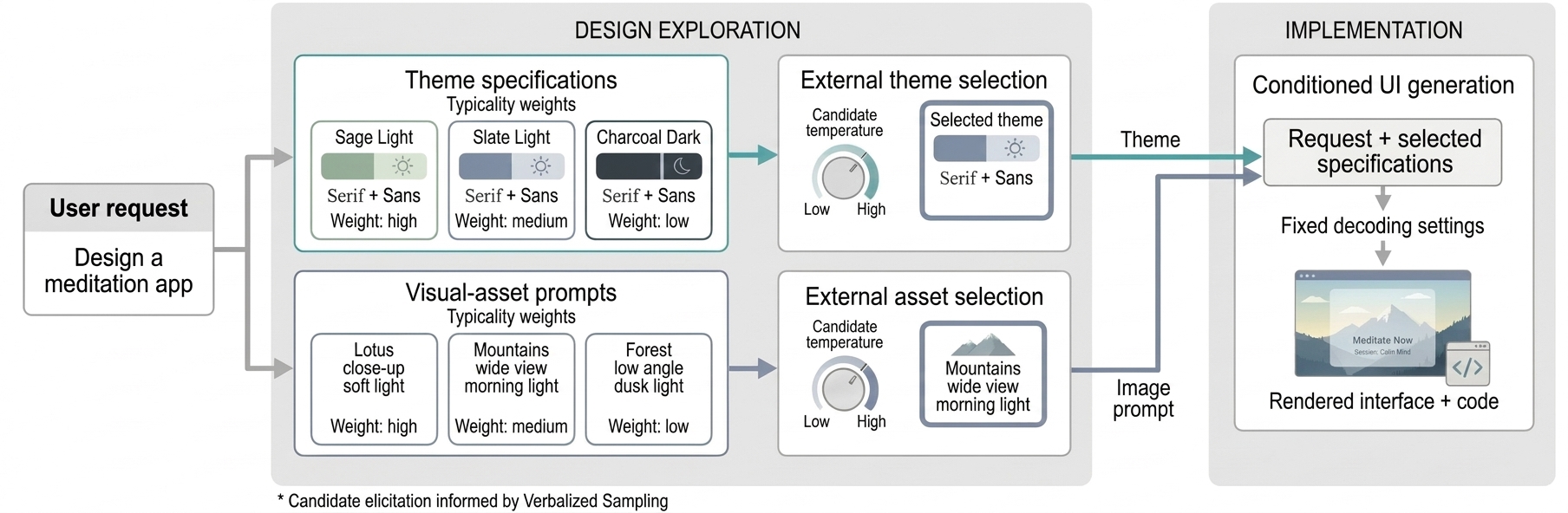}
\caption{\textbf{Design planning before UI implementation.} Before generating code, the agent proposes weighted candidates for UI themes and image prompts, samples one from each enabled branch using its own temperature, and passes the chosen design along with the user's request to the generator under fixed decoding settings. Both branches are shown together; either can run independently.}
\label{fig:teaser_pipeline}
\end{paperfigure}

Multimodal models are rapidly becoming \emph{vibe design agents}: tools where users describe an interface in plain language, inspect rendered screens, and iterate through conversation. Platforms like Lovable, v0, Bolt, and Replit Agent turn prompts directly into web apps \citep{lovable2026platform,vercel2026v0,bolt2026builder,replit2026agent}, while Figma Make, Claude Design, and Google Stitch focus on editable prototypes and UI-to-code workflows \citep{ng2025figmamake,anthropic2026claudedesign,banks2026stitch}. Recent benchmarks track rapid gains in reproducing layouts and building functional interfaces \citep{si2025design2code,xiao2025designbench,zhang2024frontend}. Yet strong coding accuracy is only half of early design work: users also want to see distinct visual directions before committing to one \citep{zamfirescu2025beyond}.

In practice, repeated requests to a UI agent often look strikingly alike, collapsing onto the same default palettes, fonts, and compositions. Post-training methods such as RLHF and DPO improve instruction following \citep{ouyang2022training,rafailov2023direct}, but preference tuning tends to narrow output diversity and suppress minority tastes \citep{xiao2025algorithmic,lanchantin2025diverse}. Verbalized Sampling points to a related cause: models inherit a typicality bias from training data that favors safe, conventional answers \citep{zhang2025verbalized}. Whatever the mix of causes, raising token-level decoding temperature is a blunt fix. Higher temperature perturbs layout, syntax, and styling all at once, and in narrative tasks it correlates only weakly with novelty while hurting coherence \citep{peeperkorn2024temperature}.

We instead pull aesthetic choices into a dedicated planning step before code generation (Figure~\ref{fig:teaser_pipeline}). Drawing on Verbalized Sampling \citep{zhang2025verbalized}, a pre-pass asks the model for structured design options (coordinated UI themes or complete image prompts) together with typicality weights. An external sampler draws one option at a branch-specific temperature, and the generator implements it alongside the user's original brief while downstream decoding stays fixed. Because expert evaluators tend to agree on usability and adherence while disagreeing on visual taste \citep{hopkins2026human}, no single score captures whether this works. We therefore track option coverage, screenshot and DOM variation, LLM-judge preferences, and live multi-turn user interaction separately.

\paragraph{Contributions and findings.}
We present an inference-time planning pipeline that gives vibe design agents explicit, separate controls over UI themes and generated imagery without touching code-generation decoding. Across 168 benchmark prompts, theme sampling substantially broadens the visual directions explored, while image-prompt sampling achieves stronger LLM-judge win rates (up to a 1.32 win/loss ratio) despite minimal shifts in whole-screen similarity. Optimal temperatures also differ across branches and prompt complexity, demonstrating why separate controls matter. Finally, a 54-day randomized production study covering 3.39 million tasks across a generator upgrade connects design exploration to real user behavior: correction rates stay flat and thumbs-down ratings trend lower (though with wide uncertainty), while the extra planning step introduces measurable latency and server-error costs.

\section{Separating Design Planning from Implementation}
\label{sec:method}

To explore visual alternatives without destabilizing UI code generation, we split inference into three stages: proposing structured design options, sampling one option externally, and conditioning the UI generator on that choice. The approach draws on Verbalized Sampling \citep{zhang2025verbalized} to elicit weighted alternatives, then grounds those candidates in concrete UI decisions before any code or imagery is produced. Downstream model parameters and token decoding settings stay identical across all conditions.

\subsection{Proposing Structured Design Options}

Given a user's UI prompt, a pre-pass asks the model for $K$ compatible design directions, each paired with a short rationale and a score reflecting how conventional (typical) the direction is. Unlike narrative outlines in hierarchical text generation \citep{yao2019planwrite}, each candidate commits concrete visual parameters for the downstream generator. Figure~\ref{fig:teaser_pipeline} shows the two branches, which can run separately or together:
\begin{itemize}[leftmargin=*,nosep]
    \item \textbf{Theme branch:} Each option specifies a coordinated bundle of five attributes: seed color, light or dark mode, headline font, body font, and corner radius. For example, a meditation-app prompt might yield a conventional light sage theme alongside a sharp-cornered dark slate alternative. Committing the bundle together prevents clashing random combinations during design-system generation.
    \item \textbf{Image-prompt branch:} Each option is a complete text-to-image prompt defining subject, composition, and lighting (for example, a close-up mason jar of overnight oats versus an editorial shot on light oak; Appendix~\ref{app:joint_case_study}), chosen before invoking the image generator.
\end{itemize}
The pipeline is agnostic to the candidate count $K$ and the specific labels used to prompt distinct tiers.

Because LLM-assigned scores need not match calibrated model probabilities \citep{hu2023prompting,wang2024calibrating}, we treat them purely as uncalibrated selection weights. Nonnegative scores are normalized to weights $\hat p_k$. Under the specified validation rules, malformed or all-zero scores trigger a retry, falling back to standard unconditioned generation if the retry also fails (the archived offline logs do not record how often retries occurred).

\subsection{Temperature-Controlled Selection}

Each enabled branch passes its normalized weights $\hat p_k > 0$ to an external sampler with its own candidate temperature $\tau > 0$:
\begin{equation}
q_k(\tau)=\frac{\hat p_k^{1/\tau}}{\sum_{j=1}^{K}\hat p_j^{1/\tau}},
\qquad \tau>0.
\label{eq:selection}
\end{equation}
At $\tau=1$, options are drawn in proportion to their normalized weights. Lower values concentrate sampling on the most conventional option; higher values flatten the distribution and give unusual directions a larger share, while zero-weight options are never chosen. Crucially, $\tau$ shifts only which design option is selected, leaving token-level decoding untouched. The chosen theme guides design-system generation, the chosen image prompt drives asset rendering, and when both branches are active, both feed into UI code generation alongside the user's original prompt.

\section{Experimental Setup}
\label{sec:eval}

We evaluate the pipeline in two settings: controlled offline benchmarks measuring option coverage, visual and DOM variation, and LLM-judge preference\ifusegoogle\footnote{\textit{AutoRater} is the internal name of the LLM-based evaluator used for pairwise UI assessments.}\fi{}, followed by a 54-day randomized A/B test in a live UI-generation product spanning a generator upgrade.

\subsection{Models and Sampling Settings}

The pre-pass requests $K=3$ labeled tiers: \textbf{\textit{Safe}} (conventional), \textbf{\textit{Premium}} (visually refined), and \textbf{\textit{Experimental}} (unexpected). These labels prompt diverse options rather than guaranteeing calibrated quality levels. In offline tests, one fast multimodal model handles candidate proposal, design-system generation, and UI code generation; a fixed text-to-image model renders images; and a larger multimodal LLM judges output pairs (archived offline exports do not record exact checkpoint versions). The online study runs on the platform's live model routing, which upgraded midway through the experiment (Section~\ref{subsec:online_setup}).

We sweep five candidate temperatures, $\tau \in \{0.5, 1, 1.5, 2, 5\}$, as a descriptive exploration rather than held-out tuning. Offline runs test the theme and image-prompt branches independently. Both the online A/B test and the qualitative case studies enable the two branches jointly at the shipped defaults ($\tau_{\text{Theme}}=1.5$, $\tau_{\text{Image}}=2.0$), which differ from the highest-scoring offline settings reported in Section~\ref{sec:results}.

\subsection{Offline Benchmark Suites}
\label{subsec:benchmarks}

We evaluate on 168 UI prompts split into two suites that test open-ended versus tightly constrained briefs:
\begin{itemize}[leftmargin=*,nosep]
    \item \textbf{Standard suite (83 prompts):} Short, open-ended requests that leave visual direction to the agent. Each prompt runs five times at both mobile and desktop viewports ($415$ pairs per viewport; $830$ pairs per $\tau$ and branch).
    \item \textbf{Complex suite (85 prompts):} Detailed briefs specifying layout hierarchy, style, and domain components ($43$ mobile and $42$ desktop prompts, five repeats each; $425$ pairs per $\tau$ and branch).
\end{itemize}
Each pair matches a baseline run against a sampled run on the same prompt and viewport using the same underlying generator, totaling $1{,}255$ pairwise comparisons per $\tau$ and branch (not $1{,}255$ unique outputs across the sweep). Tables refer to the candidate-sampling pipeline as \emph{Sampling}.

\subsection{Offline Metrics}

We track four offline measurements:
\begin{itemize}[leftmargin=*,nosep]
    \item \textbf{Option coverage:} On the $83$ standard prompts, how many of the three tier labels are chosen across five repeats, reported as mean unique options and the share of prompts stuck on a single option.
    \item \textbf{Visual variation:} Pairwise cosine similarity of screenshot embeddings across repeats of the same prompt (lower means greater visual variety).
    \item \textbf{Structural variation:} HTML DOM similarity across repeated runs of the same prompt.
    \item \textbf{LLM-judge preference:} Pairwise wins, losses, ties, and win/loss ratios across all $830$ standard and $425$ complex pairs per $\tau$, using a customized rubric with unrated pairs kept in the denominator.
\end{itemize}
These metrics measure different aspects of output behavior rather than overall design quality: distinct option labels do not guarantee visually distinct renders, similarity metrics do not verify functional code, and judge win rates reflect model rather than human taste.

\subsection{Online A/B Setup and Metrics}
\label{subsec:online_setup}

In the live product study, authenticated users are hashed by a persistent pseudonymous ID into $1{,}000$ buckets at session creation. After an initial four-day ramp at 1\% per arm, 100 buckets (10\%) receive the baseline control and 100 buckets (10\%) receive joint sampling. Within the platform, a user works in a \emph{project}, each visit is a \emph{session}, the dialogue inside a project is a \emph{conversation}, each user prompt creates a \emph{task}, and a finished task produces one or more \emph{screens}.

The product exposes a default generation mode plus optional speed and quality modes. Before the mid-study upgrade, default mode used a fast model for most steps and a larger model for selected design and editing steps. After the upgrade, default mode switched to a single newer fast model for every step (with a lighter model in speed mode). The candidate-proposal prompt and sampling temperatures ($\tau_{\text{Theme}}=1.5$, $\tau_{\text{Image}}=2.0$) remained unchanged across the upgrade.

We evaluate production impact across two sets of telemetry metrics:
\begin{itemize}[leftmargin=*,nosep]
    \item \textbf{Speed, reliability, and usage:} Task completion and failure rates; shares of tasks finishing within 60\,s, within 100\,s, or taking over 180\,s; backend server exceptions; code and design-tool exports per generated screen; and daily active users (users completing at least one task per day, summed over days).
    \item \textbf{User feedback and correction effort:} Voluntary thumbs-up and thumbs-down events per $1{,}000$ screens; an LLM-assigned correction rate (and moderate-or-worse subset) flagging follow-up turns that ask the agent to fix or redo an output (unvalidated against human labels); and \emph{single-session projects} (projects with only one session, assigned no favorable direction because one visit can indicate either immediate success or early abandonment).
\end{itemize}

\section{Results}
\label{sec:results}

We first isolate theme sampling and image-prompt sampling on offline benchmarks, then evaluate both branches together in production across a generator upgrade. Section~\ref{sec:joint_case_study} inspects rendered screens from joint sampling directly.

\subsection{Theme Sampling Broadens Visual Exploration}
\label{subsec:theme_results}

Because each theme bundles color, typography, light/dark mode, and corner radius, sampling themes shifts the look of the entire screen. On the 83 standard prompts, baseline runs pick a single option ($1.00$ on average), whereas theme sampling expands repeated runs to nearly all three proposed options (Table~\ref{tab:theme-results}). By $\tau=1.5$, zero prompts collapse onto a single option across five repeats, and screenshot similarity drops steadily from $0.6765$ (baseline) to $0.5438$ at $\tau=2.0$, showing that broader option coverage translates into visibly different screens.

Higher temperatures do not help indefinitely. At $\tau=2.0$, theme sampling pairs broad coverage ($2.87$ options) with its highest LLM-judge win/loss ratio of $1.10$ ($38.8\%$ wins vs.\ $35.3\%$ losses). Pushing to $\tau=5.0$ adds almost no visual variety ($0.5437$ vs.\ $0.5438$ similarity) while pulling the win/loss ratio back down to $0.90$. Once the sampler covers most of the proposed options, up-weighting unusual tiers further hurts appeal without adding much diversity.

\begin{papertable}[t]
\centering
\caption{Theme sampling on the standard benchmark ($83$ prompts for coverage; $N=830$ pairs per $\tau$ for judge preference). Baseline coverage is $1.00$ option and screenshot similarity is $0.6765$ (lower means greater visual variety). Unrated pairs ($0.5$--$1.1\%$) stay in the denominators. Bold marks the highest win/loss row.}
\label{tab:theme-results}
\ifuseacl\small\else\footnotesize\fi
\setlength{\tabcolsep}{4.5pt}
\renewcommand{\arraystretch}{1.06}
\begin{tabular}{@{}lrrrrrrrr@{}}
\toprule
$\tau$ & \textbf{Options} & \textbf{One option} & \textbf{Sim.} & \textbf{Diff.} & \textbf{Sampling} & \textbf{Base Win} & \textbf{Tie} & \textbf{W/L} \\
\midrule
0.5 & 2.10 & 12.0\% & 0.6560 & $-0.0204$ & 33.6\% & 41.3\% & 24.6\% & 0.81 \\
1.0 & 2.59 & 3.6\% & 0.6085 & $-0.0680$ & 36.1\% & 38.8\% & 24.2\% & 0.93 \\
1.5 & 2.84 & 0.0\% & 0.5844 & $-0.0920$ & 34.1\% & 40.1\% & 24.7\% & 0.85 \\
\keycell{2.0} & \keycell{2.87} & \keycell{0.0\%} & \keycell{0.5438} & \keycell{$-0.1327$} & \keycell{38.8\%} & \keycell{35.3\%} & \keycell{25.4\%} & \keycell{\textbf{1.10}} \\
5.0 & 2.94 & 0.0\% & 0.5437 & $-0.1328$ & 35.1\% & 38.8\% & 25.7\% & 0.90 \\
\bottomrule
\end{tabular}
\end{papertable}

\subsection{Image-Prompt Sampling Drives Judge Preference}
\label{subsec:asset_results}

\ifuseacl\begin{table}[t]\else\begin{papertable}[t]\fi
\centering
\caption{Image-prompt sampling (with theme sampling off) on the standard benchmark ($N=830$ pairs per $\tau$; baseline similarity $0.6765$). Unrated pairs ($0.0$--$0.2\%$) stay in denominators. Bold marks the highest win/loss row.}
\label{tab:asset-results}
\footnotesize
\setlength{\tabcolsep}{3.3pt}
\renewcommand{\arraystretch}{1.06}
\begin{tabular}{@{}lrrrrrr@{}}
\toprule
$\tau$ & \textbf{Sim.} & \textbf{Diff.} & \textbf{Sampling} & \textbf{Base} & \textbf{Tie} & \textbf{W/L} \\
\midrule
0.5 & 0.6588 & $-0.0177$ & 40.1\% & 33.0\% & 26.7\% & 1.22 \\
\keycell{1.0} & \keycell{0.6616} & \keycell{$-0.0148$} & \keycell{42.8\%} & \keycell{32.3\%} & \keycell{24.9\%} & \keycell{\textbf{1.32}} \\
1.5 & 0.6665 & $-0.0100$ & 33.1\% & 38.8\% & 28.1\% & 0.85 \\
2.0 & 0.6720 & $-0.0045$ & 35.2\% & 36.6\% & 28.2\% & 0.96 \\
5.0 & 0.6726 & $-0.0039$ & 34.1\% & 38.9\% & 26.7\% & 0.88 \\
\bottomrule
\end{tabular}
\ifuseacl\end{table}\else\end{papertable}\fi

Sampling image prompts alone behaves quite differently from sampling themes (Table~\ref{tab:asset-results}). Whole-screen similarity barely moves from the baseline ($-0.0148$ at $\tau=1.0$), yet judge preference jumps sharply: at $\tau=1.0$, image sampling wins $42.8\%$ of comparisons against $32.3\%$ for the baseline, a $1.32$ win/loss ratio. Even modest imagery changes inside hero cards or banners carry heavy weight with the judge when the rest of the interface stays similar.

The temperature sweet spot is also inverted relative to themes. Image sampling beats the baseline at conservative temperatures ($\tau=0.5$ and $1.0$), but falls below $1.0$ at every higher setting ($\tau \ge 1.5$). Pushing imagery too far toward experimental prompts appears to produce less appealing results, which is why the theme and image branches need separate temperatures.

\subsection{Best Temperatures Depend on Prompt Detail and Design Level}
\label{subsec:complexity_results}

Open-ended prompts leave most styling choices to the agent, whereas the 85 complex prompts pin down layout, components, and tone. Under those tighter constraints, the best image-sampling temperature shifts upward from $\tau=1.0$ to $\tau=1.5$, reaching a win/loss ratio of about $1.19$. Paired with the gap between themes ($\tau=2.0$) and images ($\tau=1.0$) on standard prompts, no single temperature works best across branches or prompt styles.

Comparing screenshots with HTML DOM trees clarifies what theme sampling actually alters: screenshot similarity drops steeply while DOM similarity stays relatively high, indicating that repeated runs vary in palette, typography, and styling while reusing similar page structures. (Shared DOM structure does not confirm functional equivalence, which requires direct interactive testing.) Judge preferences also differ between desktop and mobile runs, but because the complex suite uses different prompts for each viewport, we cannot separate screen-size effects from prompt differences.

\begin{papertable}[t]
\centering
\caption{Online A/B evaluation (10\% control vs.\ 10\% joint sampling at $\tau_{\text{Theme}}=1.5$, $\tau_{\text{Image}}=2.0$) over the full 54-day window and the stable periods before and after the generator upgrade. Initial 1\% ramp and rollout days are excluded from the before/after columns\ifuseacl{} (exact windows in Appendix~\ref{app:reproducibility})\fi. $\Delta\%$ is the relative change of sampling versus control with the platform's 95\% confidence interval; $^{\ast}$ marks intervals excluding zero. Arrows mark favorable directions. Bold highlights key feedback, correction, latency, and error rows (not significance). Before/after columns omit raw sampling counts, which equal control scaled by $\Delta\%$.}
\label{tab:online-ab-models}
\ifuseneurips\tiny\setlength{\tabcolsep}{0.5pt}\else\scriptsize\setlength{\tabcolsep}{1.4pt}\fi
\renewcommand{\arraystretch}{1.08}
\begin{tabular}{@{}l rrr rr rr@{}}
\toprule
 & \multicolumn{3}{c}{\textbf{Full study}} & \multicolumn{2}{c}{\textbf{Before upgrade}} & \multicolumn{2}{c}{\textbf{After upgrade}} \\
\cmidrule(lr){2-4}\cmidrule(lr){5-6}\cmidrule(l){7-8}
\textbf{Metric} & \textbf{Control} & \textbf{Sampling} & \textbf{$\Delta\%$ [95\% CI]} & \textbf{Control} & \textbf{$\Delta\%$ [95\% CI]} & \textbf{Control} & \textbf{$\Delta\%$ [95\% CI]} \\
\midrule
Created tasks & 1,700,939 & 1,691,119 & $-0.6$ $[-2.7, +1.6]$ & 346,464 & $-0.3$ $[-3.8, +3.2]$ & 1,021,393 & $-1.0$ $[-3.2, +1.3]$ \\
Generated screens & 2,185,029 & 2,178,334 & $-0.3$ $[-2.7, +2.1]$ & 575,115 & $-0.4$ $[-3.3, +2.6]$ & 1,180,994 & $-0.1$ $[-3.0, +2.8]$ \\
LLM-labeled conversations & 44,018 & 43,398 & $-1.4$ $[-3.4, +0.6]$ & 9,719 & $-1.3$ $[-5.6, +2.9]$ & 26,628 & $-1.7$ $[-3.8, +0.3]$ \\
Daily active users (sum) & 247,399 & 244,249 & $-1.3^{\ast}$ $[-2.5, -0.1]$ & 55,399 & $-0.1$ $[-2.2, +2.0]$ & 142,082 & $-1.6^{\ast}$ $[-3.0, -0.2]$ \\
\midrule
\keycell{Thumbs-down / 1K screens $\downarrow$} & 0.197 & 0.168 & \keycell{$-14.6$ $[-32.8, +3.5]$} & 0.270 & $-22.3$ $[-50.6, +6.0]$ & 0.171 & $-10.3$ $[-33.7, +13.1]$ \\
Thumbs-down count $\downarrow$ & 430 & 366 & $-14.9$ $[-33.5, +3.7]$ & 155 & $-22.6$ $[-51.0, +5.9]$ & 202 & $-10.4$ $[-34.1, +13.3]$ \\
\keycell{Correction rate $\downarrow$} & 37.91\% & 38.05\% & \keycell{$+0.4$ $[-2.0, +2.7]$} & 39.78\% & $+0.4$ $[-2.4, +3.2]$ & 37.57\% & $+0.9$ $[-2.3, +4.1]$ \\
Moderate-or-worse corrections $\downarrow$ & 29.00\% & 28.57\% & $-1.5$ $[-4.4, +1.4]$ & 31.35\% & $-0.9$ $[-4.1, +2.4]$ & 28.43\% & $-0.8$ $[-4.3, +2.8]$ \\
Single-session projects & 40.62\% & 41.03\% & $+1.0$ $[-3.6, +5.6]$ & 43.10\% & $+1.6$ $[-2.8, +6.1]$ & 39.61\% & $+0.4$ $[-4.9, +5.7]$ \\
\midrule
Task completion rate $\uparrow$ & 97.39\% & 97.21\% & $-0.18$ $[-0.53, +0.16]$ & 98.01\% & $-0.090^{\ast}$ $[-0.177, -0.002]$ & 97.77\% & $-0.25$ $[-0.76, +0.27]$ \\
Task failure rate $\downarrow$ & 1.75\% & 1.91\% & $+9.2$ $[-10.5, +29.0]$ & 1.35\% & $+2.8$ $[-3.0, +8.6]$ & 1.26\% & $+17.9$ $[-25.0, +60.7]$ \\
Completed within 60\,s $\uparrow$ & 51.61\% & 51.10\% & $-1.0$ $[-2.7, +0.7]$ & 47.41\% & $-0.9$ $[-2.4, +0.5]$ & 53.20\% & $-1.1$ $[-3.5, +1.2]$ \\
\keycell{Completed within 100\,s $\uparrow$} & 71.71\% & 70.80\% & \keycell{$-1.3^{\ast}$ $[-2.2, -0.3]$} & 73.73\% & $-2.2^{\ast}$ $[-2.9, -1.5]$ & 71.36\% & $-1.1$ $[-2.4, +0.1]$ \\
Longer than 180\,s $\downarrow$ & 8.90\% & 9.17\% & $+3.1$ $[-0.9, +7.0]$ & 6.10\% & $+4.3$ $[-2.0, +10.6]$ & 10.06\% & $+3.3$ $[-0.6, +7.1]$ \\
\keycell{Server exception rate $\downarrow$} & 0.762\% & 0.799\% & \keycell{$+4.9^{\ast}$ $[+0.1, +9.6]$} & 1.303\% & $+4.1$ $[-1.8, +10.1]$ & 0.328\% & $+15.6^{\ast}$ $[+5.4, +25.9]$ \\
Code export / screen $\uparrow$ & 1.41\% & 1.34\% & $-5.0$ $[-14.5, +4.5]$ & 0.99\% & $-1.3$ $[-17.1, +14.4]$ & 1.61\% & $-6.6$ $[-17.5, +4.2]$ \\
Design-tool export / screen $\uparrow$ & 4.44\% & 4.29\% & $-3.5$ $[-12.5, +5.6]$ & 3.15\% & $-0.9$ $[-10.6, +8.8]$ & 5.11\% & $-5.4$ $[-16.0, +5.1]$ \\
\bottomrule
\end{tabular}
\end{papertable}

\subsection{Joint Sampling in Production}
\label{subsec:interactive_friction}

Over 54 days, the live A/B study logged 3.39 million tasks and 719,882 projects with balanced request traffic across arms (Table~\ref{tab:online-ab-models}; confidence intervals are the platform's unadjusted bucket estimates\ifuseacl{}, see Appendix~\ref{app:reproducibility}\fi). A primary risk of bolder visual choices is that users might ask the agent to fix more screens. Across 87,416 LLM-labeled conversations, that did not materialize: correction rates stayed virtually flat at $37.91\%$ under control versus $38.05\%$ under joint sampling ($+0.36\%$, 95\% CI $[-2.01\%, +2.73\%]$), and neither moderate-or-worse corrections nor single-session project rates moved appreciably.

Voluntary ratings point in a mildly positive direction, though with wide uncertainty. Joint sampling saw $14.6\%$ fewer thumbs-down ratings per 1,000 screens (95\% CI $[-32.8\%, +3.5\%]$), trending lower in six of eight weekly cohorts and under both generator setups. Still, the confidence interval crosses zero, only 6,112 voluntary ratings were submitted overall, and neither code exports nor design-tool exports improved.

The clearest production effects are operational costs. Adding the planning pass reduced the share of tasks finishing within 100\,s from $71.71\%$ to $70.80\%$ ($-1.3\%$) and lifted server exceptions from $0.762\%$ to $0.799\%$ ($+4.9\%$); both intervals exclude zero. While small in absolute terms, those regressions compound at million-task scale, and an analysis of the final 19 days shows larger exception spikes, lower task completion, and more runs exceeding 180\,s\ifuseacl{} (Appendix~\ref{app:reproducibility})\fi. Summed daily active users were also $1.3\%$ lower overall and $1.6\%$ lower after the upgrade, though diagnosing whether this reflects churn rather than imbalance requires user-level assignment and retention logs. Sparse client-side browser error logs likewise leave frontend reliability unverified.

\subsection{Stability Across a Generator Upgrade}
\label{subsec:model_upgrade}

Midway through the study, the platform replaced its multi-model default routing with a single newer fast model. Comparing sampling against its concurrent control during the stable windows before and after the rollout (Table~\ref{tab:online-ab-models}) reveals how much larger the platform upgrade was than the sampling effect. Control-arm server exceptions dropped from $1.303\%$ to $0.328\%$, tasks finishing within 60\,s rose from $47.41\%$ to $53.20\%$, and export rates climbed markedly. Because traffic and seasonality also shift over time, those before-versus-after jumps are not causal estimates of the upgrade, which underscores why a concurrent control is indispensable when evaluating inference interventions in production.

Within each period, user-facing patterns stayed consistent: correction rates remained flat, thumbs-down point estimates stayed negative (with CIs crossing zero), and export rates showed no gain. Operational regressions shifted in relative visibility, with the within-100\,s latency drop clearest before the upgrade ($-2.2\%$) and the relative server-exception jump sharpest afterward ($+15.6\%$ off a much lower $0.328\%$ baseline), though confirming whether treatment effects truly changed requires a direct cross-period test. Because the proposal prompt and temperatures ($\tau_{\text{Theme}}=1.5$, $\tau_{\text{Image}}=2.0$) were never retuned for the newer model, how much recalibration would recover remains open (\ifuseacl Section~\ref{sec:discussion}\else Section~\ref{sec:lessons}\fi).

\ifuseacl
\begin{paperfigure}[t]
\centering
\ifdefined\ACLFinal\def\caseStudyPanelWidth{0.36\textwidth}\else\def\caseStudyPanelWidth{0.43\textwidth}\fi
\textbf{\small (a) Baseline:} {\footnotesize light sage-green palettes; \texttt{Newsreader} headings.}\\[0.15em]
\begin{minipage}[t]{\caseStudyPanelWidth}
  \vspace{0pt}\centering
  \includegraphics[width=\linewidth,trim=34bp 1258bp 34bp 0,clip]{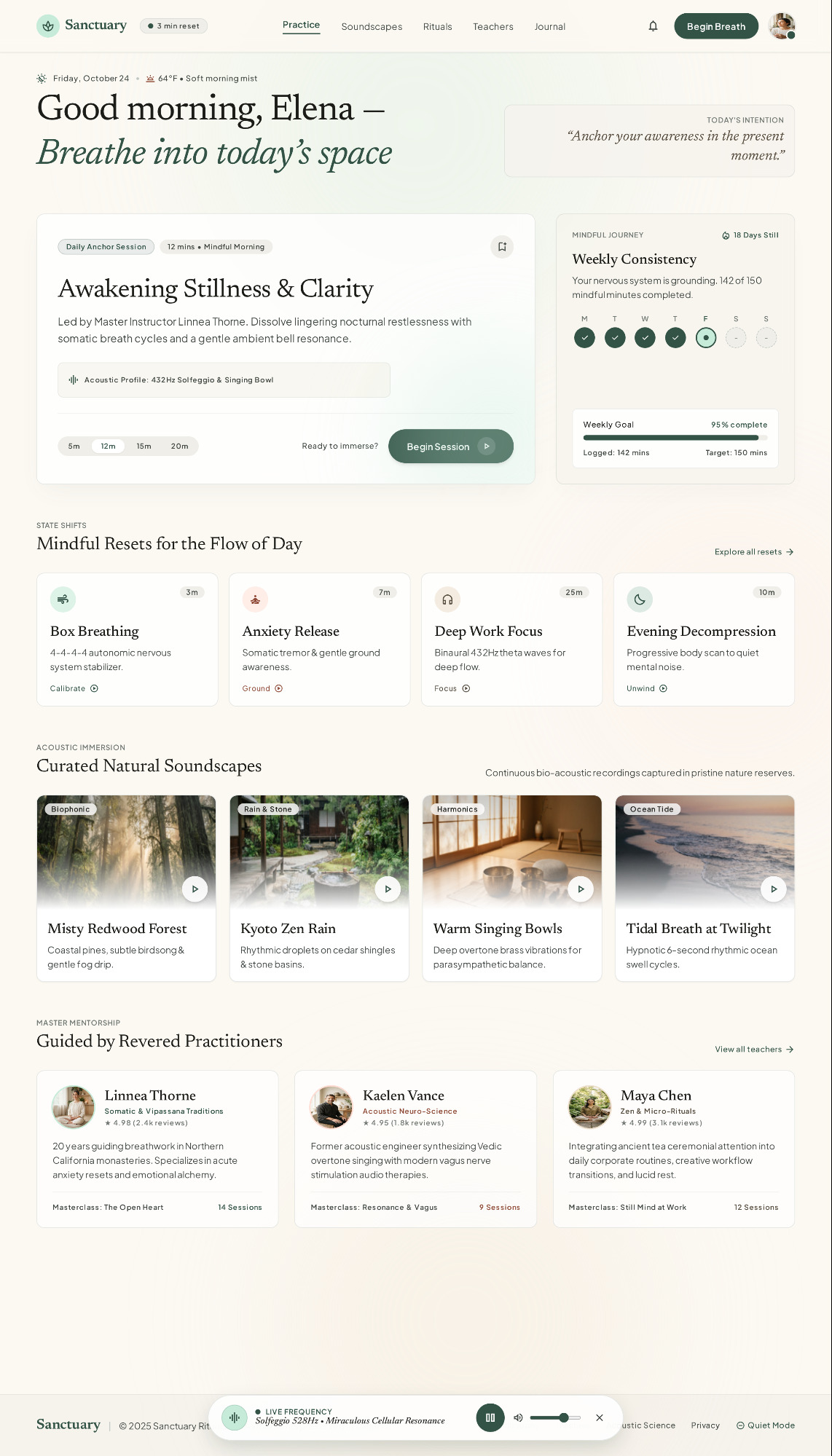}\\[-0.1em]
  {\scriptsize Run 1 (\texttt{\#4a6b5d}, light)}
\end{minipage}\hspace{0.02\textwidth}
\begin{minipage}[t]{\caseStudyPanelWidth}
  \vspace{0pt}\centering
  \includegraphics[width=\linewidth,trim=39bp 1154bp 39bp 0,clip]{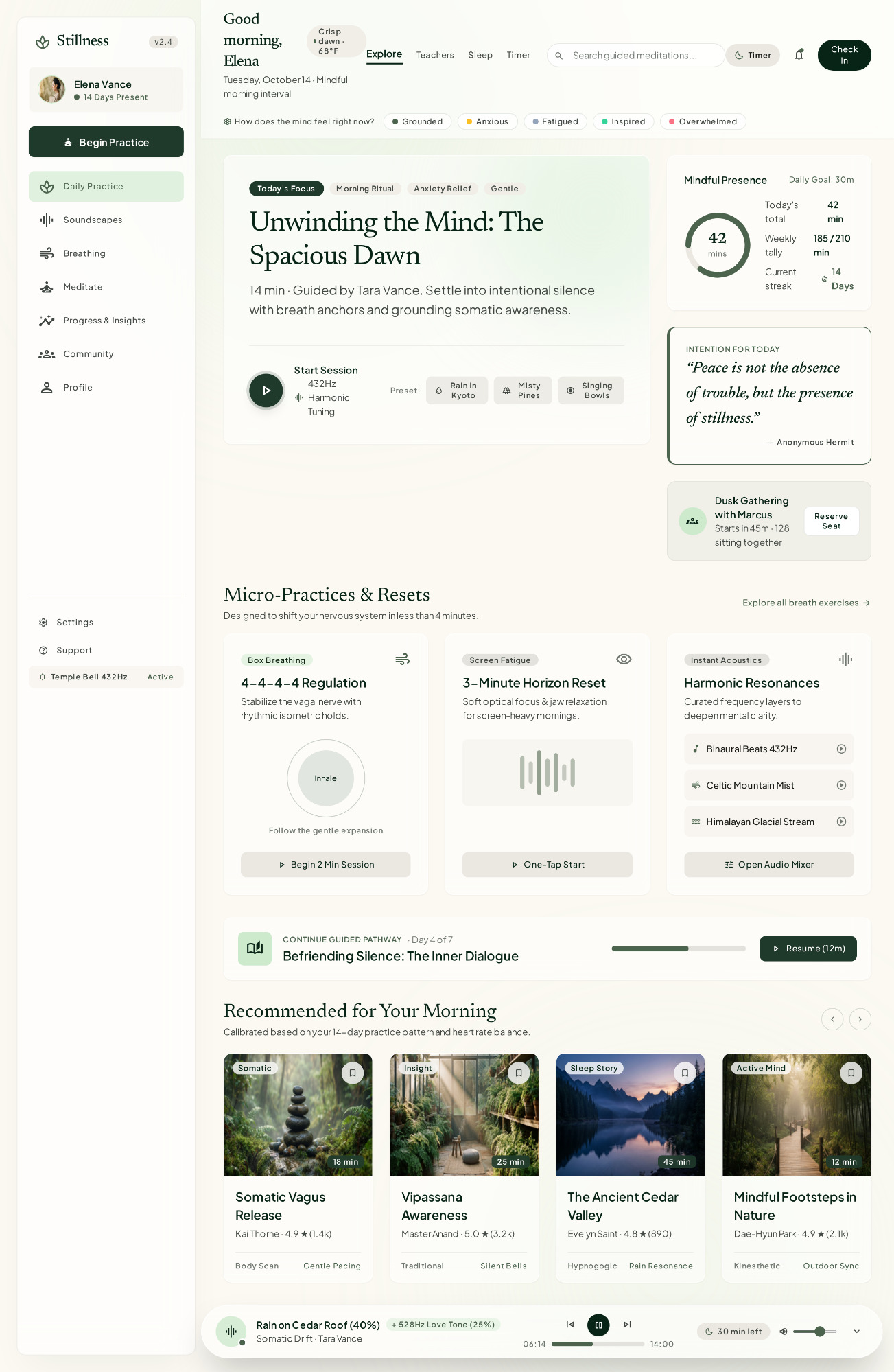}\\[-0.1em]
  {\scriptsize Run 2 (\texttt{\#1f3a2b}, light)}
\end{minipage}\\[0.45em]
\textbf{\small (b) Joint candidate sampling:} {\footnotesize slate-eucalyptus and dark nocturnal themes; varied serifs.}\\[0.15em]
\begin{minipage}[t]{\caseStudyPanelWidth}
  \vspace{0pt}\centering
  \includegraphics[width=\linewidth,trim=39bp 1154bp 39bp 0,clip]{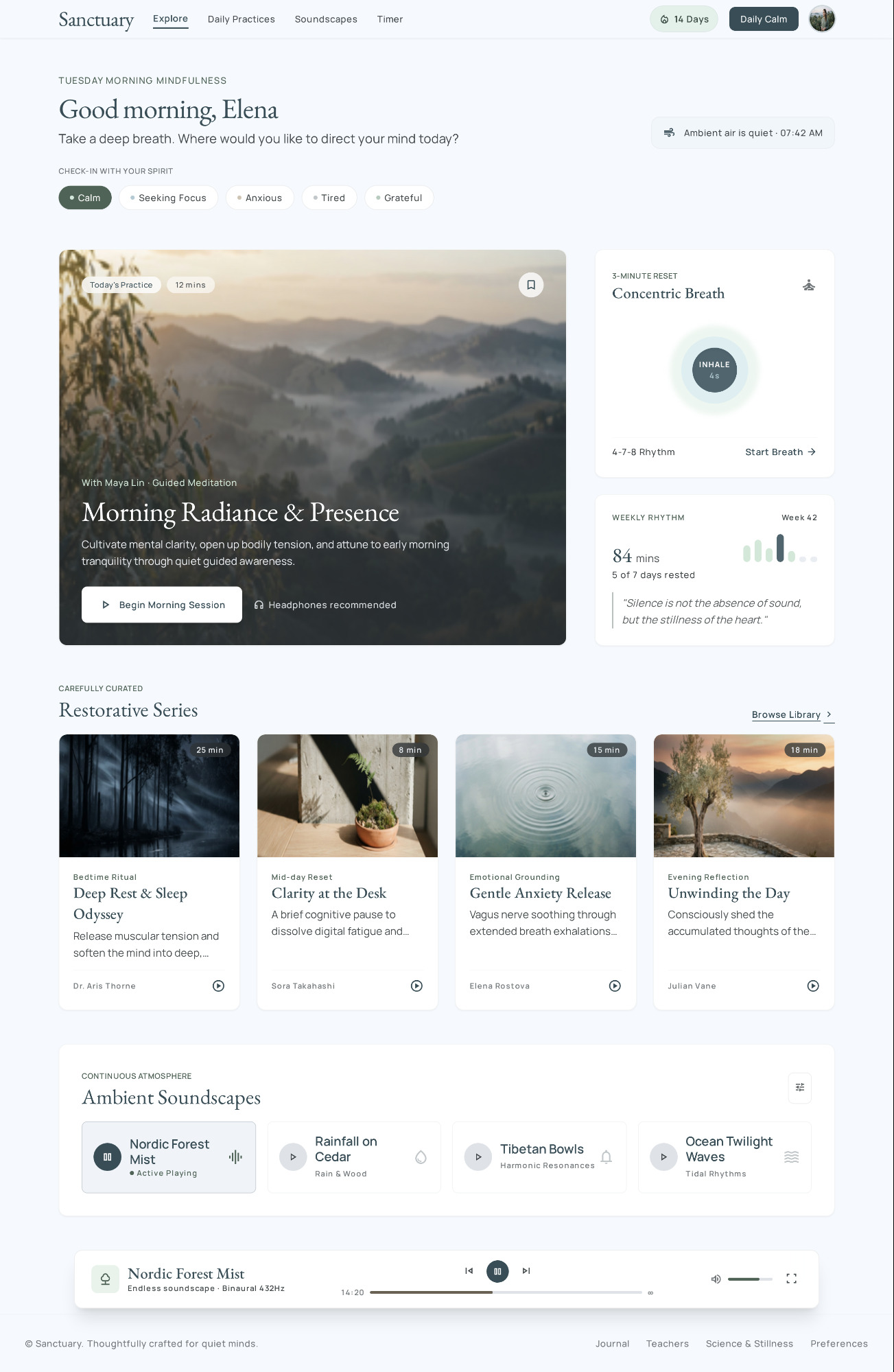}\\[-0.1em]
  {\scriptsize Run 2 (\textit{Premium}, $\hat p=0.25$)}
\end{minipage}\hspace{0.02\textwidth}
\begin{minipage}[t]{\caseStudyPanelWidth}
  \vspace{0pt}\centering
  \includegraphics[width=\linewidth,trim=28bp 1387bp 28bp 0,clip]{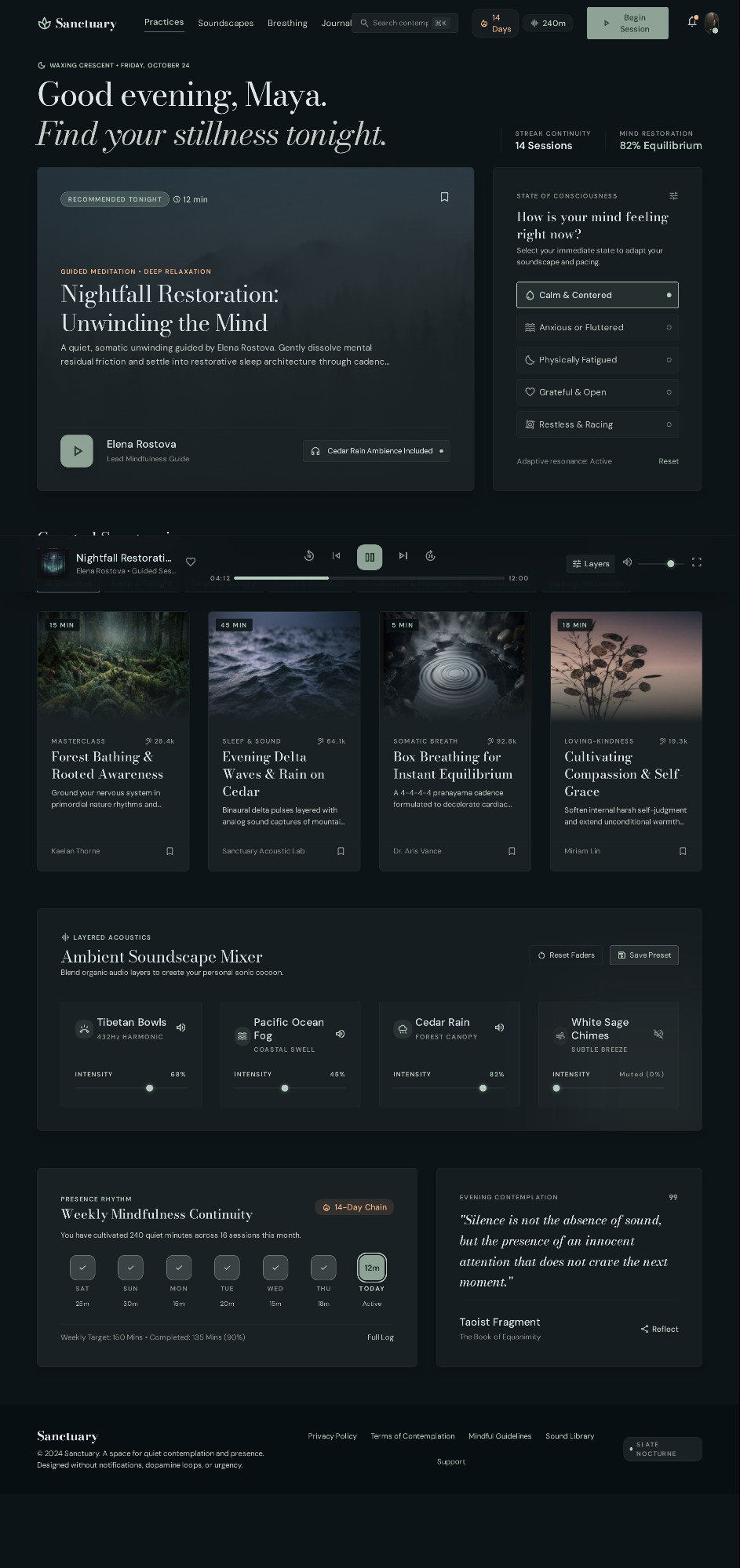}\\[-0.1em]
  {\scriptsize Run 3 (\textit{Experimental}, $\hat p=0.10$)}
\end{minipage}
\caption{Enlarged upper views of meditation-app generations for the same brief (1280px viewport), with two runs per condition. Full five-run comparisons, design-token matrices, and a mobile case study are in Appendix~\ref{app:joint_case_study}.}
\label{fig:meditation_comparison}
\end{paperfigure}

\else
\begin{paperfigure}[t]
\ifuseacl\def\caseStudyImageHeight{3.8cm}\else\def\caseStudyImageHeight{4.6cm}\fi
\centering
\begin{minipage}[t]{0.485\textwidth}
  \vspace{0pt}
  \centering
  \textbf{\small (a) Baseline}\\[0.2em]
  \begin{minipage}[t]{0.485\linewidth}
    \vspace{0pt}
  \centering
    \begin{minipage}[b][\caseStudyImageHeight][b]{\linewidth}
    \centering
    \includegraphics[width=\linewidth,height=\caseStudyImageHeight,keepaspectratio,trim=34bp 925bp 34bp 0,clip]{assets/optimized/meditation_no_vs/screen_1.jpg}
    \end{minipage}\\[-0.1em]
    {\scriptsize Run 1 (\texttt{\#4a6b5d}, light)}
  \end{minipage}\hfill
  \begin{minipage}[t]{0.485\linewidth}
    \vspace{0pt}
  \centering
    \begin{minipage}[b][\caseStudyImageHeight][b]{\linewidth}
    \centering
    \includegraphics[width=\linewidth,height=\caseStudyImageHeight,keepaspectratio,trim=39bp 774bp 39bp 0,clip]{assets/optimized/meditation_no_vs/screen_2.jpg}
    \end{minipage}\\[-0.1em]
    {\scriptsize Run 2 (\texttt{\#1f3a2b}, light)}
  \end{minipage}\\[0.2em]
  {\footnotesize Light sage-green palettes; \texttt{Newsreader} headings.}
\end{minipage}\hfill
\begin{minipage}[t]{0.485\textwidth}
  \vspace{0pt}
  \centering
  \textbf{\small (b) Joint candidate sampling}\\[0.2em]
  \begin{minipage}[t]{0.485\linewidth}
    \vspace{0pt}
  \centering
    \begin{minipage}[b][\caseStudyImageHeight][b]{\linewidth}
    \centering
    \includegraphics[width=\linewidth,height=\caseStudyImageHeight,keepaspectratio,trim=39bp 774bp 39bp 0,clip]{assets/optimized/meditation_vs/screen_2.jpg}
    \end{minipage}\\[-0.1em]
    {\scriptsize Run 2 (\textit{Premium}, $\hat p=0.25$)}
  \end{minipage}\hfill
  \begin{minipage}[t]{0.485\linewidth}
    \vspace{0pt}
  \centering
    \begin{minipage}[b][\caseStudyImageHeight][b]{\linewidth}
    \centering
    \includegraphics[width=\linewidth,height=\caseStudyImageHeight,keepaspectratio,trim=28bp 1112bp 28bp 0,clip]{assets/optimized/meditation_vs/screen_3.jpg}
    \end{minipage}\\[-0.1em]
    {\scriptsize Run 3 (\textit{Experimental}, $\hat p=0.10$)}
  \end{minipage}\\[0.2em]
  {\footnotesize Slate-eucalyptus and dark nocturnal themes; varied serifs.}
\end{minipage}
\caption{Cropped upper views of desktop meditation-app generations (\textit{``Design the home screen for a meditation app''}, 1280px viewport), showing two runs per condition. Full screenshots of all five runs per condition, design-token matrices, and a mobile case study appear in Appendix~\ref{app:joint_case_study}.}
\label{fig:meditation_comparison}
\end{paperfigure}

\fi

\section{Qualitative Case Study: Joint Theme and Image Sampling}
\label{sec:joint_case_study}

The offline benchmarks isolate theme and image sampling one branch at a time. To inspect how they work together on rendered screens, we ran five baseline and five joint-sampling generations on two prompts: a desktop meditation home screen (Figure~\ref{fig:meditation_comparison}) and a mobile meal-planning dashboard (all ten screenshots and extracted tokens appear in Appendix~\ref{app:joint_case_study}).

Without sampling, all five meditation runs converge on the same look: light sage-green palettes, \texttt{Newsreader} headings, and sunlit nature scenes. Joint sampling keeps that familiar style on some runs while introducing clear departures, including a muted slate-eucalyptus layout and two dark-mode screens with sharper corners and high-contrast serifs (Figure~\ref{fig:meditation_comparison}b). The mobile meal-planning runs show the same pattern on a denser interface: baseline repeats the same green palette in 4 of 5 runs and reuses the exact same headline font and corner radius in all 5, whereas joint sampling produces five distinct palettes, three headline fonts, three corner-radius styles, and varied food photography (Tables~\ref{tab:token-diversity}--\ref{tab:image-prompt-comparison}). These ten-run comparisons illustrate what joint sampling changes visually, though judging human preference or interactive correctness requires dedicated testing.

\ifuseacl
  \input{sections/05_acl_discussion.tex}
\else
  \section{Limitations and Implications}
\label{sec:lessons}

\subsection{What the Experiments Isolate (and Do Not)}

Separating theme and image planning reveals two complementary levers: theme sampling drives broad visual variety, while image-prompt sampling lifts judge preference with minimal whole-screen change. Because an image choice can sway perceived appeal without shifting global screenshot embeddings, diversity and preference need separate metrics.

At the same time, our sweeps test the full proposal-plus-selection pipeline against unconditioned generation. Isolating each piece requires controlled ablations against unweighted candidate proposal, uniform selection, the original Verbalized Sampling protocol, and token-temperature sweeps on the downstream generator. Because archived offline exports do not confirm whether candidate sets were reused across temperatures or repeats, our offline tables characterize the tested configurations rather than isolating $\tau$ alone. Sampling also operates strictly within the proposed options, uses uncalibrated weights, and counts tier labels whose visual identity across runs is not independently verified. Confirming that generated screens faithfully follow both the chosen design and the user's functional brief requires direct execution checks.

\subsection{Limits of Offline Judges and Live Telemetry}

Because the best temperatures shift across branches and prompt suites, these numbers should not be transferred blindly to other domains. LLM judges scale cheaply, but they carry well-documented position, length, and alignment biases \citep{zheng2023judging,verga2024replacing}. Our archived offline aggregates also lack per-prompt records needed for prompt-clustered confidence intervals, and judge scores cannot substitute for blinded evaluations by human designers.

Live telemetry adds realism but leaves its own gaps. Over the full 54-day window, correction rates stayed flat (not reproducing an earlier 9-day interim uptick), thumbs-down estimates trended lower with a confidence interval crossing zero, exports showed no gain, and latency and server exceptions worsened. Meanwhile, the mid-study generator upgrade shifted baseline metrics far more than sampling did, proving why concurrent controls are essential. Because proposal prompts and temperatures stayed fixed throughout the upgrade, whether retuning would improve the quality-versus-latency tradeoff remains untested.

\subsection{Reproducibility and Next Steps}

The proprietary prompt sets, generated outputs, evaluator logs, renderer state, and user-level telemetry cannot be released. Although our online study spans two production generator setups within one UI agent, external replication across other model families, languages, domains, and accessibility-constrained briefs is needed to test generality. Future confirmatory studies should freeze the prompt list, candidate sets, model versions, random seeds, renderer, and exclusion rules in advance; compare unweighted, uniform, and token-temperature baselines; keep unrated and invalid runs in denominators; and compute prompt-clustered uncertainty. Following the Human Creativity Benchmark \citep{hopkins2026human}, evaluation should separate adherence and working execution from visual appeal while preserving disagreement among blinded professional designers, backed by browser task tests and automated accessibility audits.

\subsection{Practical Implications}

Treating visual exploration as an explicit pre-pass exposes a simple control knob without destabilizing code generation. In practice, exploration strength should adapt to the user's brief and workflow stage (higher during early brainstorming, tighter once brand or layout constraints are fixed) and be recalibrated whenever the underlying generator upgrades. Production deployments should also pair candidate proposal and output validation with explicit guardrails for accessibility, privacy, brand rules, and task constraints.

\fi
\section{Related Work}
\label{sec:related}

\paragraph{Post-training diversity and inference-time elicitation.}
Preference tuning with RLHF and DPO improves instruction following \citep{ouyang2022training,rafailov2023direct}, but often narrows output diversity and suppresses minority tastes \citep{xiao2025algorithmic,lanchantin2025diverse}. At inference time, multi-response prompting asks models for several alternatives in one pass or across turns \citep{troshin2025asking}, and Verbalized Sampling pairs candidates with typicality scores to surface less conventional answers \citep{zhang2025verbalized}. Because verbalized scores can diverge from next-token probabilities and are not calibrated probabilities out of the box \citep{hu2023prompting,wang2024calibrating}, we normalize them strictly as selection weights and ground each candidate in concrete UI decisions.

\paragraph{UI code generation and design-space exploration.}
Early screenshot-to-code systems translated pixels directly into UI trees \citep{beltramelli2018pix2code,wu2021screen}. Modern benchmarks evaluate visual fidelity and cross-framework generation \citep{laurenccon2024unlocking,yun2024web2code,gui2025webcode2m,si2025design2code,xiao2025designbench,sun2025fullfront,zhu2025frontendbench}, as well as multi-page navigation, interactivity, automated repair, and canvas editing \citep{xiao2025interaction2code,wan2024mrweb,yuan2025designrepair,jeong2026canvas}. Recent generators improve fidelity through hierarchical planning, layout segmentation, component reuse, and iterative refinement \citep{wu2024uicoder,wan2025divide,gui2025uicopilot,zhou2025declarui,xiao2026efficientuicoder,xiao2026comuicoder}. A parallel line of interactive tools helps designers browse alternatives via layout constraints \citep{swearngin2020scout}, visual blending or staged navigation \citep{zhang2024frontend,yuan2024maxprototyper,lu2025misty,zhong2021spacewalker}, and explicit program design choices \citep{zamfirescu2025beyond}. Our pipeline complements these systems by adding lightweight, inference-time control over UI themes and image prompts before code generation begins.

\paragraph{Evaluating creative outputs and UI preferences.}
Creative quality spans multiple dimensions that a single scalar score can obscure \citep{lee2022rethinking,treffinger1972}. On the Human Creativity Benchmark, professional evaluators agree closely on adherence and technical execution while diverging on aesthetic appeal \citep{hopkins2026human}. Annotator-disagreement frameworks preserve those valid differences rather than collapsing them into one label \citep{dawid1979maximum,inel2014,basile2021we}, while embedding-based metrics track visual diversity separately from quality \citep{ibarrola2024diversity}. Learned UI reward models and LLM judges enable large-scale pairwise scoring \citep{wu2024uiclip}, though LLM judges exhibit position, length, and self-preference biases that make human designer feedback an essential complement \citep{zheng2023judging,shi2025judging,verga2024replacing}.

\section{Conclusion}
\label{sec:conclusion}

Moving visual choices into a short planning step before code generation gives vibe design agents controllable aesthetic variety without perturbing downstream decoding. Across 168 benchmark prompts, theme sampling expands repeated generations from one repeated style to nearly three distinct options, while image-prompt sampling lifts LLM-judge preference (up to a $1.32$ win/loss ratio) despite leaving whole-screen similarity almost unchanged. In a 54-day production A/B test covering 3.39 million tasks across a generator upgrade, joint sampling held user correction rates steady and pointed toward fewer thumbs-down ratings (with wide uncertainty), while exposing concrete latency and server-error costs that future systems must budget for and retune across model updates.

\ifuseacl
  \clearpage
  \input{sections/07_acl_limitations.tex}
  \input{sections/08_acl_ethics.tex}
\fi

\ifuseneurips
  \clearpage
\fi
\ifuseacl\else
  \bibliographystyle{\paperbibliographystyle}
\fi
\ifuseacl\else
  \renewcommand{\bibsection}{}
  \section*{References}
\fi
\ifusegoogle\else
  \bibliography{references}
\fi
\ifusegoogle
  \begingroup
  \setlength{\bibsep}{2pt plus 0.3pt}
  \bibliography{references}
  \endgroup
\fi

\clearpage
\appendix
\ifuseacl
  \onecolumn
  \begingroup
  \renewenvironment{papertable}[1][t]{\begin{table}[H]}{\end{table}}
\fi
\section{Extended Qualitative Case Studies}
\label{app:joint_case_study}

This appendix shows all five baseline and five joint-sampling runs, extracted design tokens, and image prompts for both qualitative case studies from Section~\ref{sec:joint_case_study}: the desktop meditation home screen (Section~\ref{app:meditation_case_study}) and the mobile meal-planning dashboard (Section~\ref{app:mealplan_case_study}).

\subsection{Desktop Meditation App (5 vs.\ 5 Runs)}
\label{app:meditation_case_study}

We ran five independent generations per condition on \textit{``Design the home screen for a meditation app''} at a 1280px desktop viewport. Table~\ref{tab:meditation-tokens} lists the extracted design tokens, corner-radius levels (1--3 and their CSS values), typicality weights, and concept rationales across all ten runs. Figures~\ref{fig:meditation_remaining}--\ref{fig:meditation_remaining-10} display all ten full screenshots, two per page, in run order; cropped previews of baseline Runs 1--2 and sampling Runs 2--3 appear in Figure~\ref{fig:meditation_comparison}.

\begin{papertable}[tp]
\centering
\caption{Extracted design tokens, corner-radius levels, typicality weights, and visual rationales across five baseline and five joint-sampling runs for the desktop meditation prompt. Every baseline run chooses light mode, sage-green palettes, and \texttt{Newsreader} headings, whereas joint sampling spans light and dark modes, four headline serifs, and all three corner-radius levels. Bold highlights varied typography and radii; the two dark-mode runs are shaded in bold.}
\label{tab:meditation-tokens}
\scriptsize
\setlength{\tabcolsep}{4pt}
\renewcommand{\arraystretch}{1.18}
\begin{tabularx}{\textwidth}{@{}l l l l l l c l X@{}}
\toprule
\textbf{Arm} & \textbf{Run} & \textbf{Seed} & \textbf{Mode} & \textbf{Headline Font} & \textbf{Body Font} & \textbf{Radius} & \textbf{Tier ($\hat p$)} & \textbf{Concept / Visual Rationale} \\
\midrule
\multirow{5}{*}{\textbf{Baseline}}
 & Run 1 & \texttt{\#4a6b5d} & light & Newsreader & Plus Jakarta Sans & 2 & Single option & \textit{Warm Ethereal Sanctuary} (8 morning/forest/zen assets) \\
 & Run 2 & \texttt{\#1f3a2b} & light & Newsreader & Plus Jakarta Sans & 2 & Single option & \textit{Serene Sanctuary} (5 stone/botanical/lake assets) \\
 & Run 3 & \texttt{\#4d6b53} & light & Newsreader & Source Sans 3 & 2 & Single option & \textit{Aura Mind} (1 practitioner portrait asset) \\
 & Run 4 & \texttt{\#58705c} & light & Newsreader & Plus Jakarta Sans & 3 & Single option & \textit{Serene Organic Sanctuary} (12 session/avatar assets) \\
 & Run 5 & \texttt{\#5b7563} & light & Newsreader & Manrope & 2 & Single option & \textit{Serene Grounding} (daytime sage sanctuary) \\
\midrule
\multirow{5}{*}{\textbf{\shortstack[l]{Candidate\\sampling}}}
 & Run 1 & \texttt{\#6C8A74} & light & \textbf{Playfair Display} & Plus Jakarta Sans & \textbf{3} & \textit{Safe} (0.65) & \textit{Prana:} Sage green with soft curves (level 3, base radius 1\,rem). \\
 & Run 2 & \texttt{\#50656E} & light & \textbf{EB Garamond} & Manrope & 2 & \textbf{\textit{Premium}} (0.25) & \textit{Stillness Meditation:} Muted slate eucalyptus with a classical serif. \\
& \keycell{Run 3} & \keycell{\texttt{\#1E2528}} & \keycell{\textbf{dark}} & \keycell{\textbf{Bodoni Moda}} & \keycell{DM Sans} & \keycell{\textbf{1}} & \keycell{\textbf{\textit{Experimental}} (0.10)} & \textit{Nocturne Sanctuary:} Dark slate with high-contrast serif headings and sharp geometry (level 1, base radius 0.125\,rem). \\
& \keycell{Run 4} & \keycell{\texttt{\#766C8D}} & \keycell{\textbf{dark}} & \keycell{\textbf{EB Garamond}} & \keycell{Manrope} & \keycell{2} & \keycell{\textbf{\textit{Premium}} (0.25)} & \textit{Sanctuary Twilight:} Muted lavender and warm clay accents in a dark theme. \\
 & Run 5 & \texttt{\#60725C} & light & \textbf{Newsreader} & Plus Jakarta Sans & 2 & \textit{Safe} (0.65) & \textit{Serene Sanctuary:} Sage palette with serif headings and warm linen tones. \\
\bottomrule
\end{tabularx}
\vspace{1mm}
{\raggedright \scriptsize Corner-radius levels map to CSS radii (base / large / extra-large): level 1 $\rightarrow$ 0.125 / 0.25 / 0.5\,rem; level 2 $\rightarrow$ 0.25 / 0.5 / 0.75\,rem; level 3 $\rightarrow$ 1 / 2 / 3\,rem.\par}
\end{papertable}

Generated imagery shifts alongside the theme tokens in Table~\ref{tab:meditation-tokens}. Baseline runs stick mostly to morning nature scenes and instructor portraits, whereas joint sampling pairs each theme with matching imagery:
\begin{itemize}[leftmargin=*,nosep]
    \item \textbf{Run 1 (\textit{Prana}, \textit{Safe} $\hat p=0.65$):} Soft curves (corner-radius level 3, large radius 2\,rem), translucent panels, and six instructor portraits in warm morning light.
    \item \textbf{Run 2 (\textit{Stillness Meditation}, \textit{Premium} $\hat p=0.25$):} Slate-eucalyptus styling paired with five landscape and still-life images (\textit{Night forest}, \textit{Minimal desk}, \textit{Rippling water}, and \textit{Sunset mountains}).
    \item \textbf{Run 3 (\textit{Nocturne Sanctuary}, \textit{Experimental} $\hat p=0.10$):} Dark mineral-slate background (\texttt{\#1E2528}), high-contrast \texttt{Bodoni Moda} headings, sharp corners (level 1, base radius 0.125\,rem), and seven low-light images (\textit{Nightfall tranquil pine forest}, \textit{Evening Delta Waves on Cedar}, and \textit{Nightfall Restoration artwork}).
    \item \textbf{Run 4 (\textit{Sanctuary Twilight}, \textit{Premium} $\hat p=0.25$):} Restrained evening layout with muted lavender (\texttt{\#766C8D}) and warm clay accents.
    \item \textbf{Run 5 (\textit{Serene Sanctuary}, \textit{Safe} $\hat p=0.65$):} Five soundscape-themed images, including \textit{Kyoto Temple Rain}, \textit{Norwegian Fjord Dawn}, \textit{432Hz Deep Theta Healing}, and \textit{Luminous Pine Canopy}.
\end{itemize}

\clearpage
\begin{paperfigure}[p]
\centering
\sbox0{\includegraphics[width=0.485\textwidth,height=0.82\textheight,keepaspectratio]{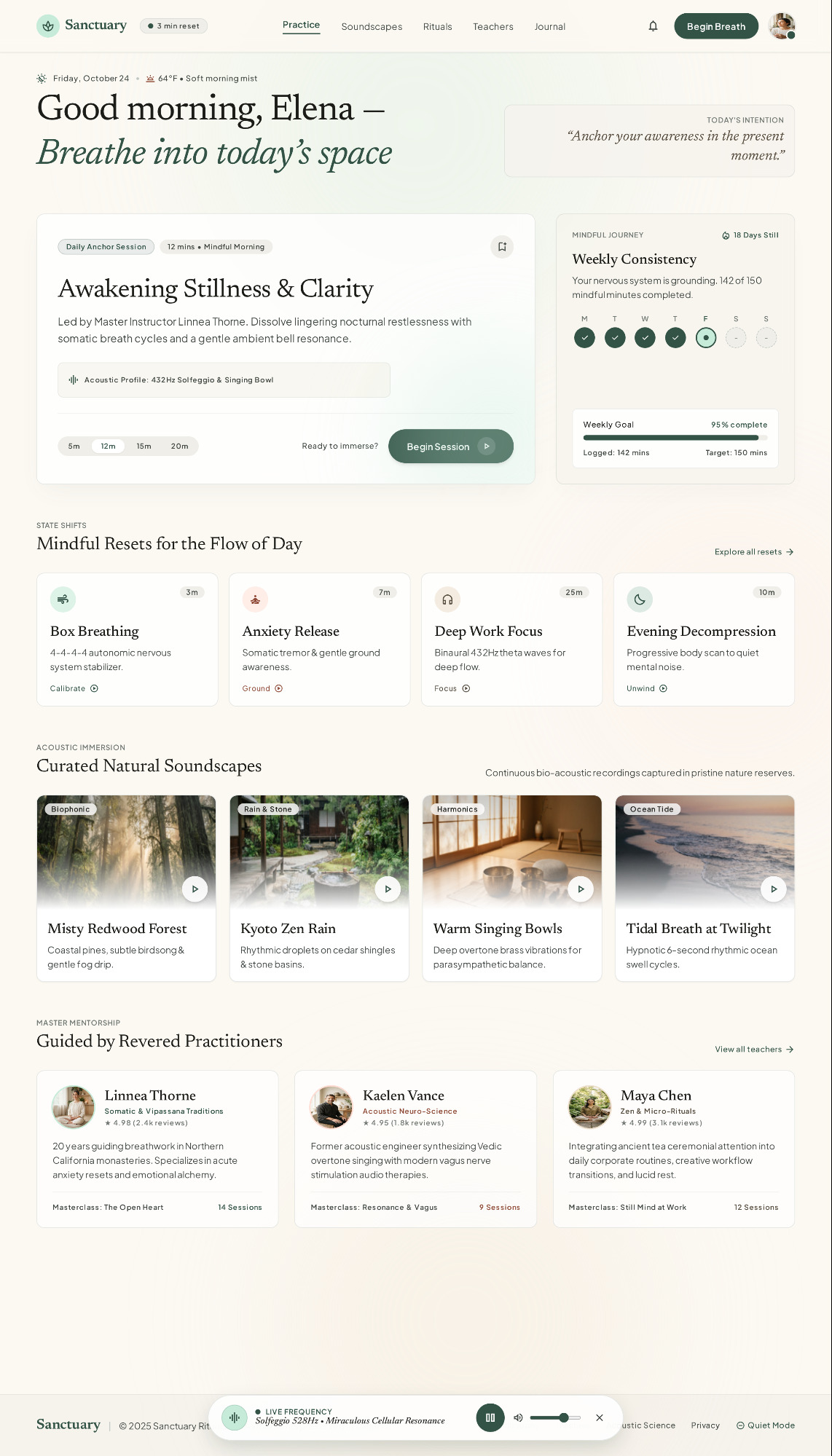}}
\sbox2{\includegraphics[width=0.485\textwidth,height=0.82\textheight,keepaspectratio]{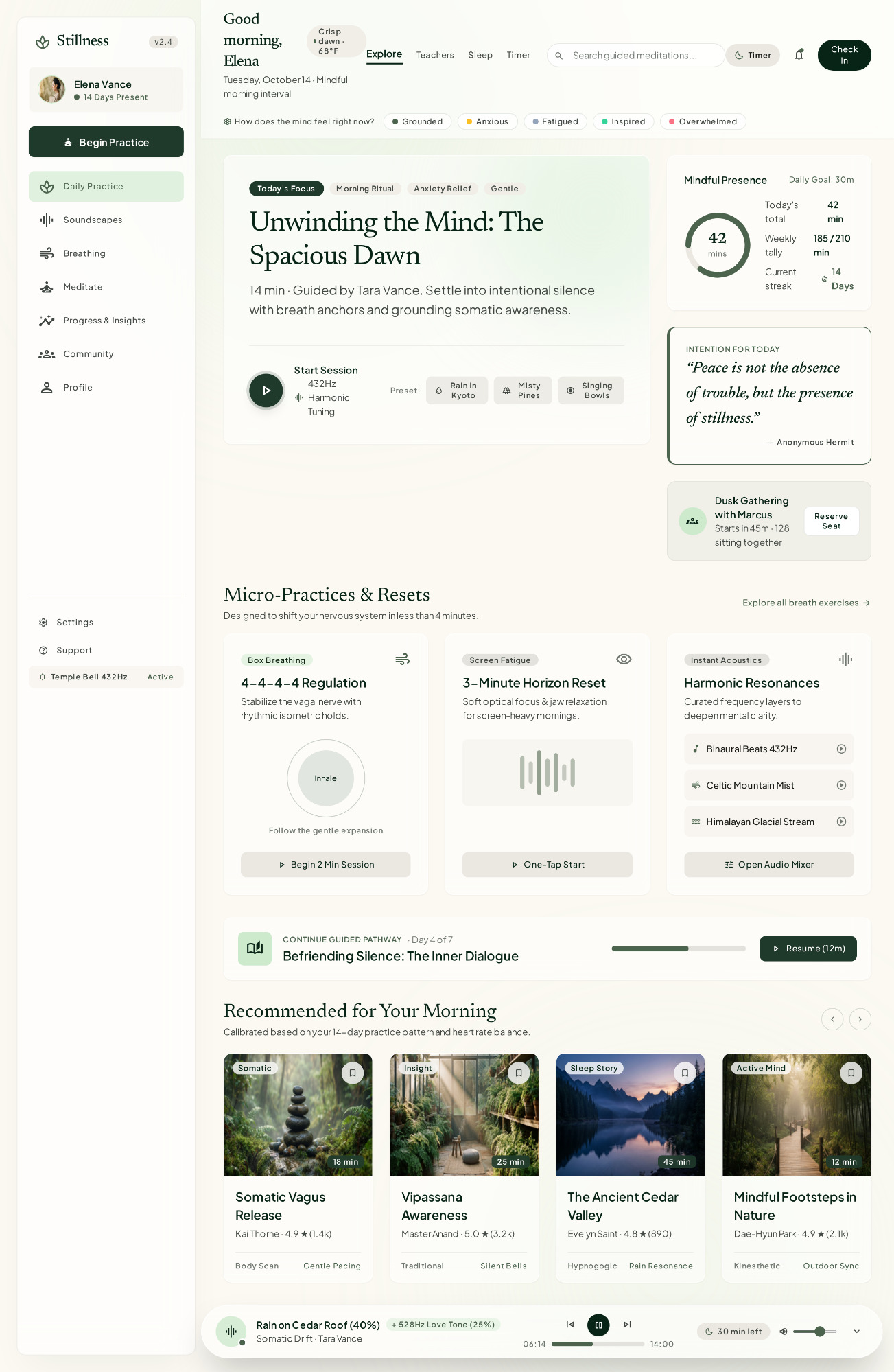}}
\dimen0=\ht0
\ifdim\ht2<\dimen0 \dimen0=\ht2\fi
\edef\appendixPairHeight{\the\dimen0}
\begin{minipage}[t]{0.485\textwidth}
\vspace{0pt}
\centering
\begin{minipage}[t][\appendixPairHeight][t]{\linewidth}
\centering
\includegraphics[height=\appendixPairHeight]{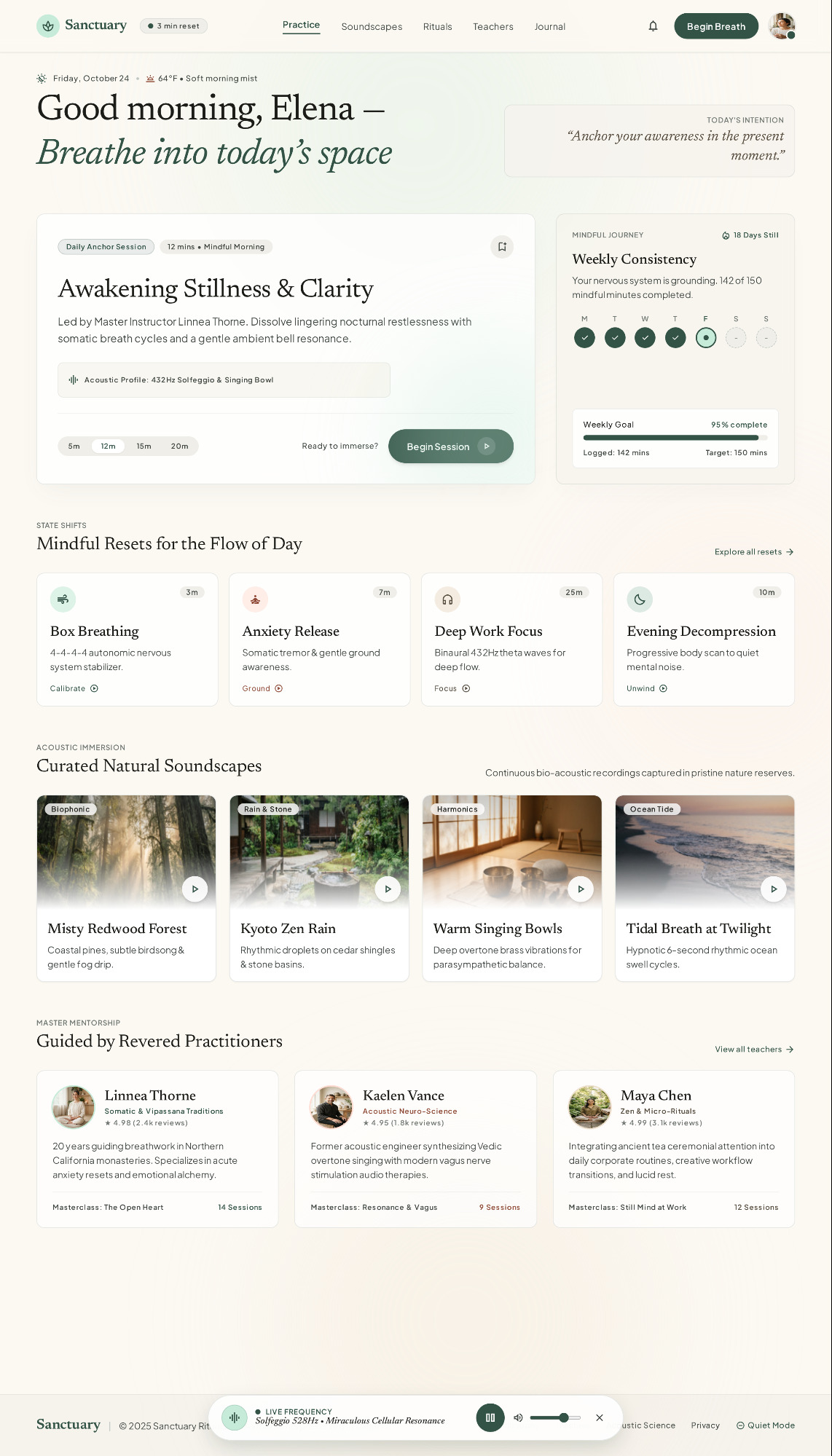}
\end{minipage}
\caption{Baseline desktop meditation-app generation: Run 1 (\texttt{\#4a6b5d}, light). Full screenshot.}
\label{fig:meditation_remaining}
\end{minipage}\hfill
\begin{minipage}[t]{0.485\textwidth}
\vspace{0pt}
\centering
\begin{minipage}[t][\appendixPairHeight][t]{\linewidth}
\centering
\includegraphics[height=\appendixPairHeight]{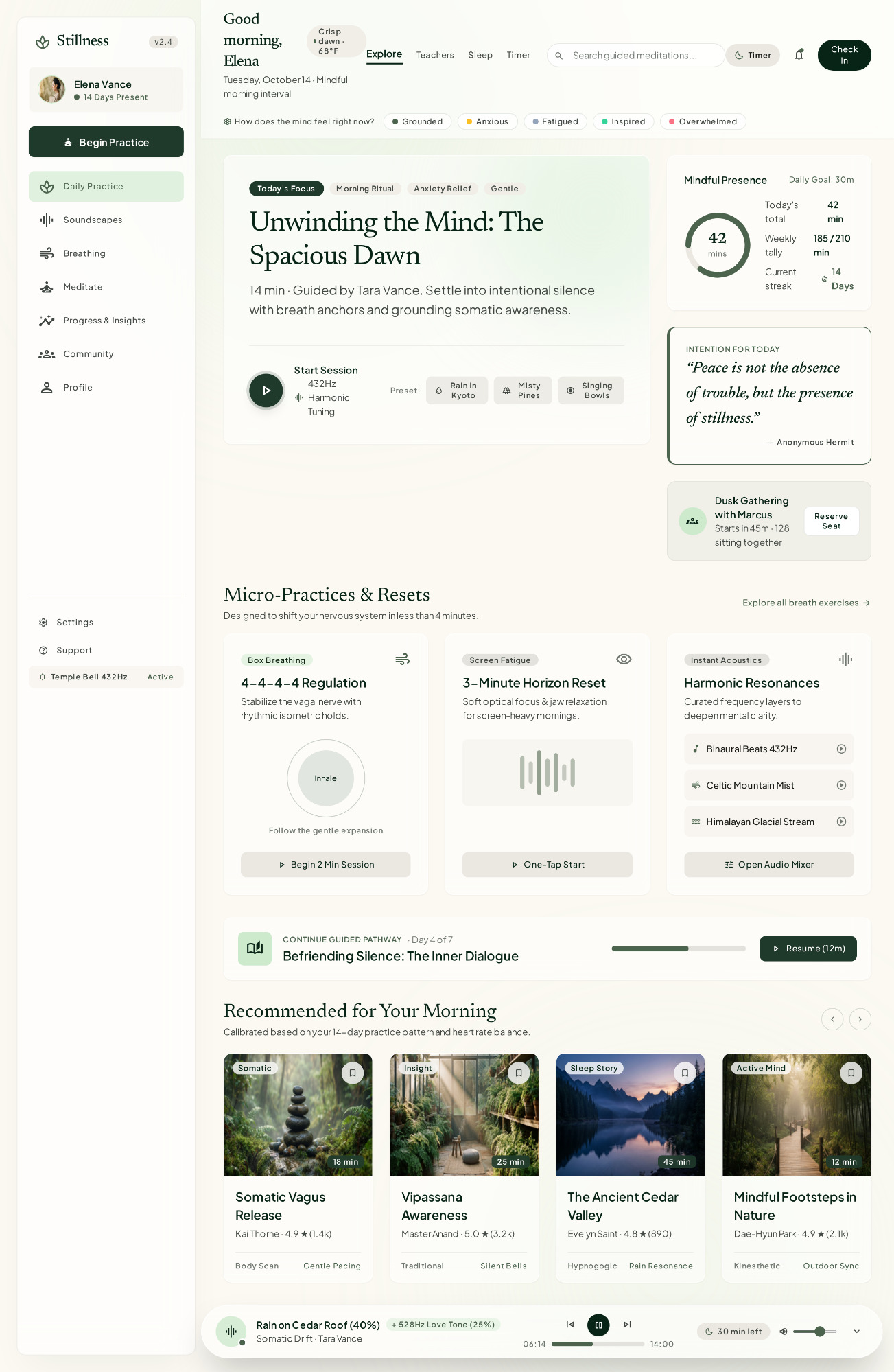}
\end{minipage}
\caption{Baseline desktop meditation-app generation: Run 2 (\texttt{\#1f3a2b}, light). Full screenshot.}
\label{fig:meditation_remaining-2}
\end{minipage}
\end{paperfigure}
\clearpage

\begin{paperfigure}[p]
\centering
\sbox0{\includegraphics[width=0.485\textwidth,height=0.82\textheight,keepaspectratio]{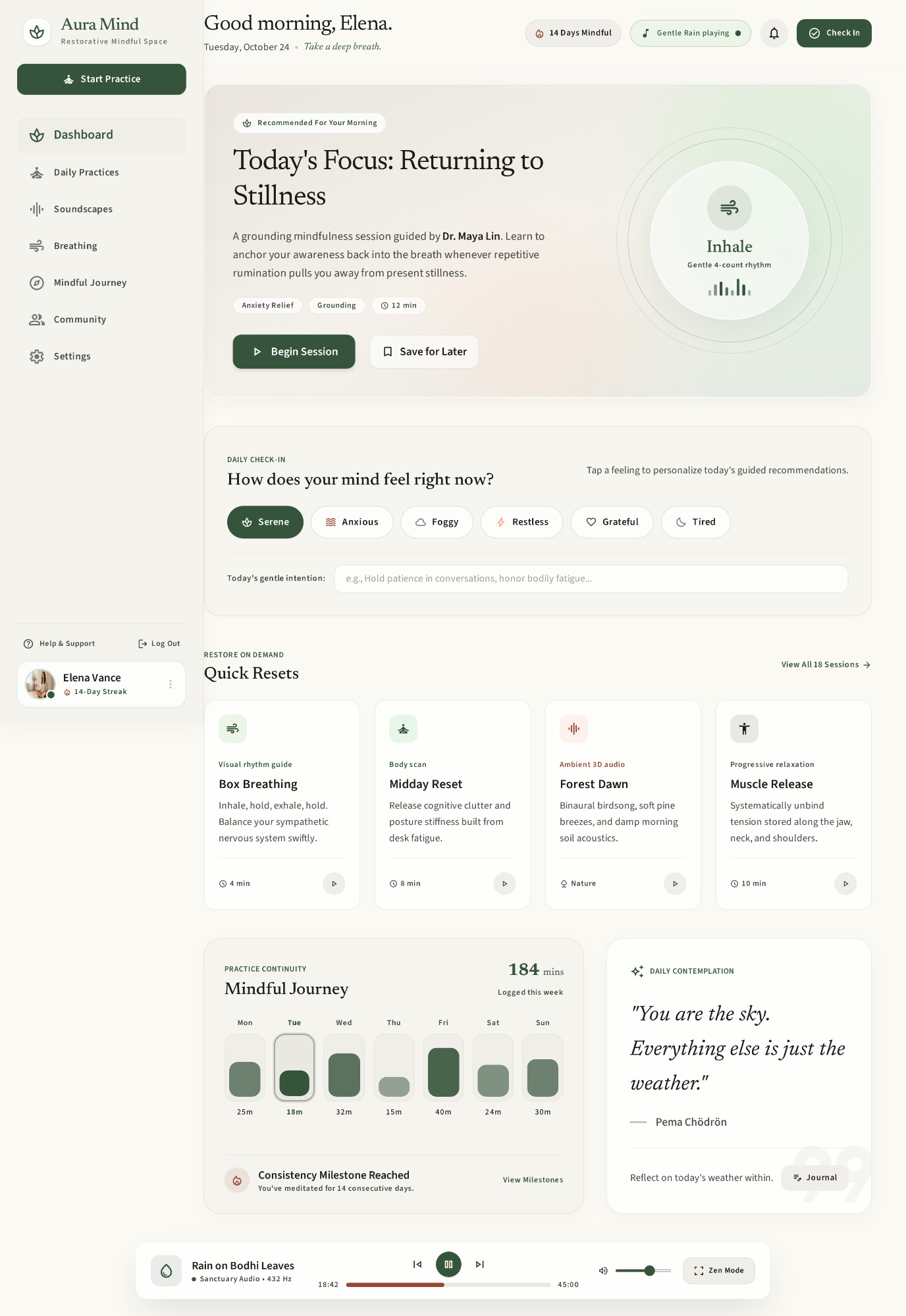}}
\sbox2{\includegraphics[width=0.485\textwidth,height=0.82\textheight,keepaspectratio]{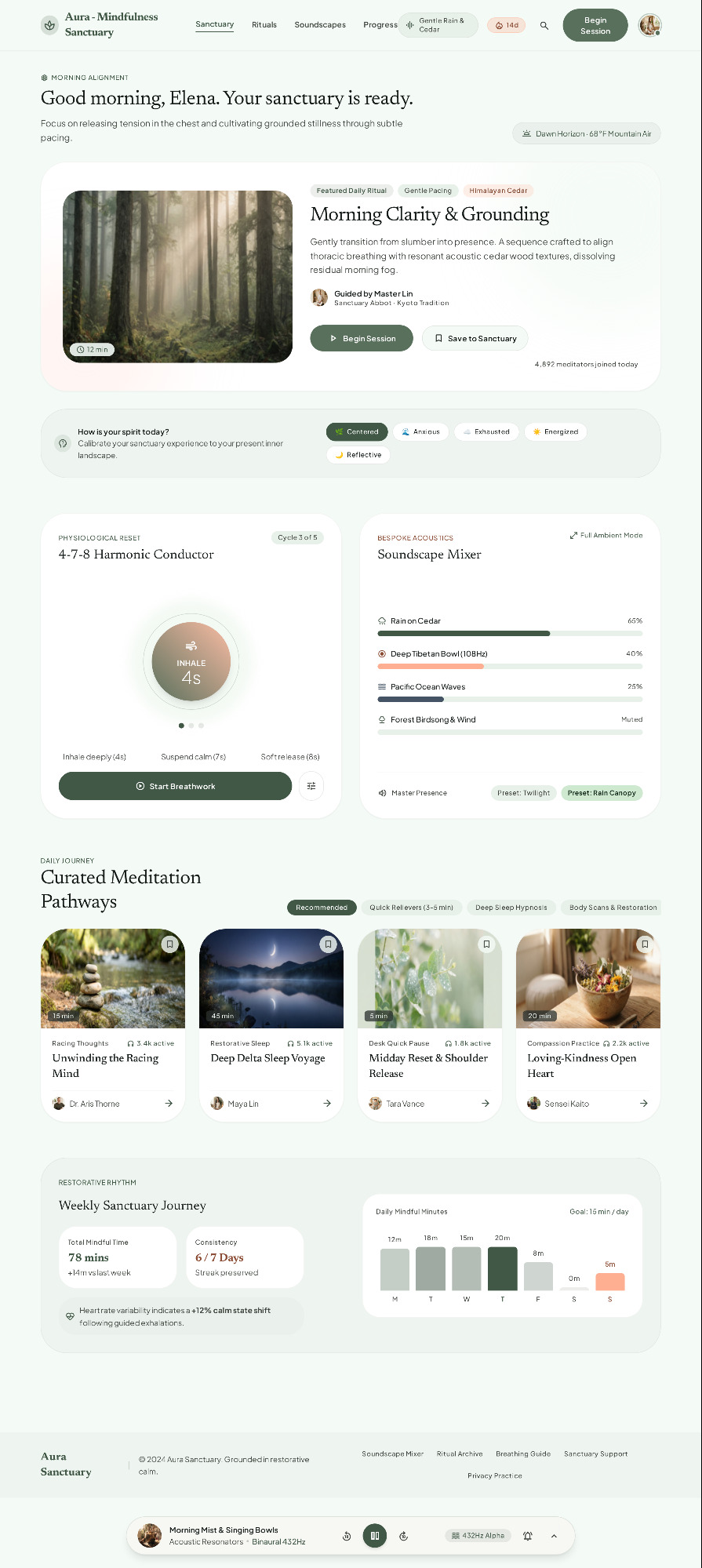}}
\dimen0=\ht0
\ifdim\ht2<\dimen0 \dimen0=\ht2\fi
\edef\appendixPairHeight{\the\dimen0}
\begin{minipage}[t]{0.485\textwidth}
\vspace{0pt}
\centering
\begin{minipage}[t][\appendixPairHeight][t]{\linewidth}
\centering
\includegraphics[height=\appendixPairHeight]{assets/optimized/meditation_no_vs/screen_3.jpg}
\end{minipage}
\caption{Baseline desktop meditation-app generation: Run 3 (\texttt{\#4d6b53}, light). Full screenshot.}
\label{fig:meditation_remaining-3}
\end{minipage}\hfill
\begin{minipage}[t]{0.485\textwidth}
\vspace{0pt}
\centering
\begin{minipage}[t][\appendixPairHeight][t]{\linewidth}
\centering
\includegraphics[height=\appendixPairHeight]{assets/optimized/meditation_no_vs/screen_4.jpg}
\end{minipage}
\caption{Baseline desktop meditation-app generation: Run 4 (\texttt{\#58705c}, light). Full screenshot.}
\label{fig:meditation_remaining-4}
\end{minipage}
\end{paperfigure}
\clearpage

\begin{paperfigure}[p]
\centering
\sbox0{\includegraphics[width=0.485\textwidth,height=0.82\textheight,keepaspectratio]{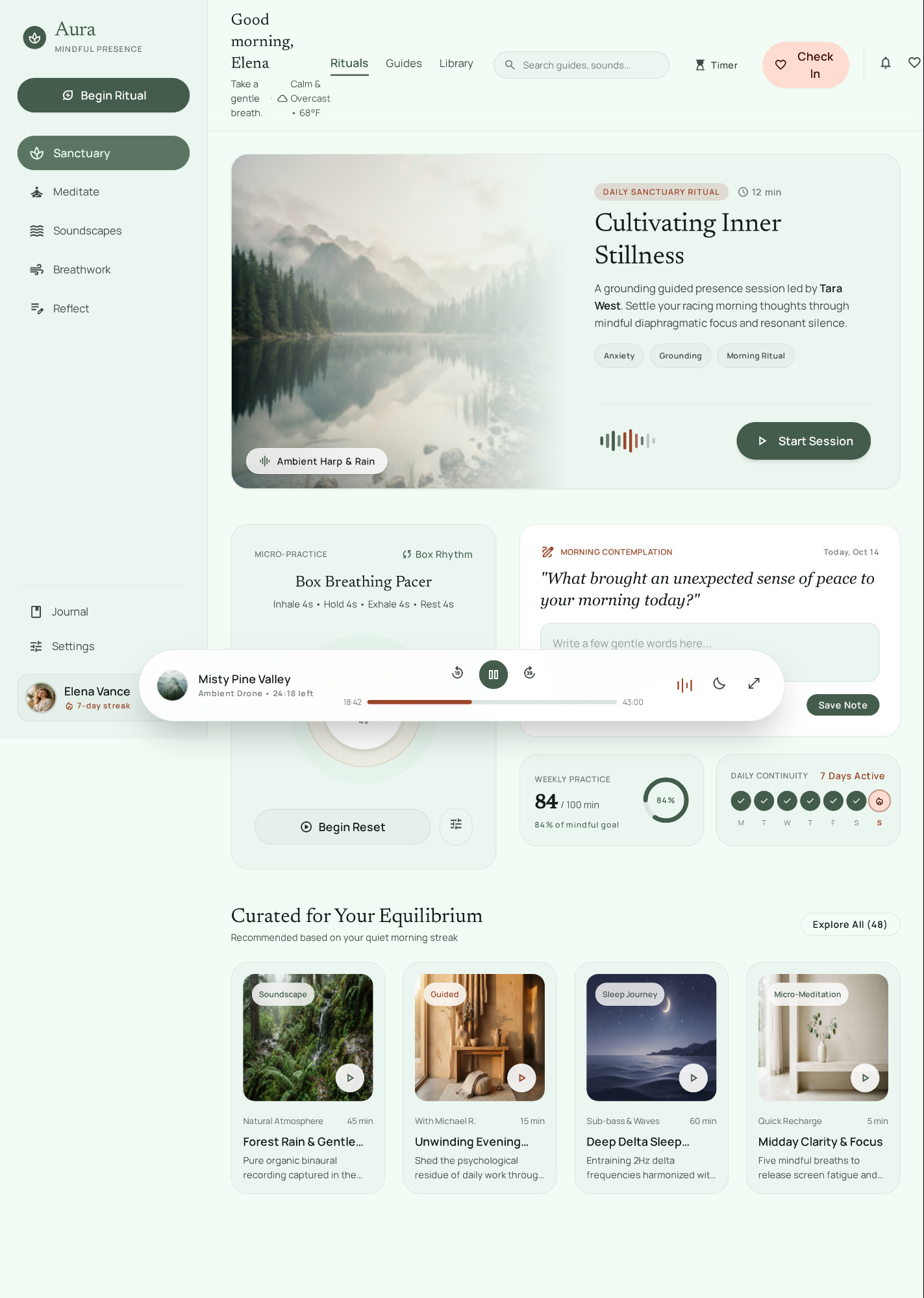}}
\sbox2{\includegraphics[width=0.485\textwidth,height=0.82\textheight,keepaspectratio]{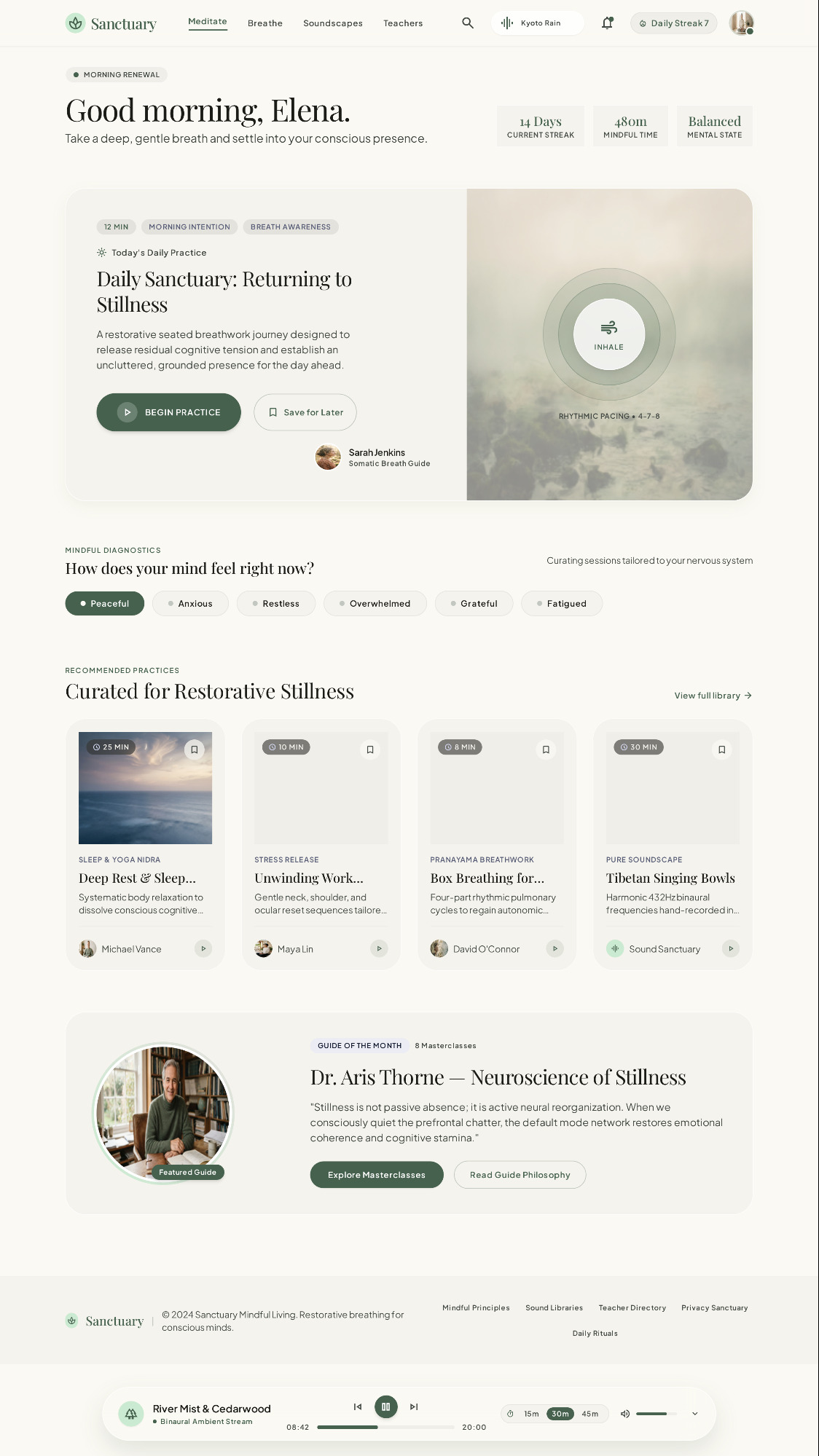}}
\dimen0=\ht0
\ifdim\ht2<\dimen0 \dimen0=\ht2\fi
\edef\appendixPairHeight{\the\dimen0}
\begin{minipage}[t]{0.485\textwidth}
\vspace{0pt}
\centering
\begin{minipage}[t][\appendixPairHeight][t]{\linewidth}
\centering
\includegraphics[height=\appendixPairHeight]{assets/optimized/meditation_no_vs/screen_5.jpg}
\end{minipage}
\caption{Baseline desktop meditation-app generation: Run 5 (\texttt{\#5b7563}, light). Full screenshot.}
\label{fig:meditation_remaining-5}
\end{minipage}\hfill
\begin{minipage}[t]{0.485\textwidth}
\vspace{0pt}
\centering
\begin{minipage}[t][\appendixPairHeight][t]{\linewidth}
\centering
\includegraphics[height=\appendixPairHeight]{assets/optimized/meditation_vs/screen_1.jpg}
\end{minipage}
\caption{Joint candidate sampling desktop meditation-app generation: Run 1 (\textit{Safe}, $0.65$, light). Full screenshot.}
\label{fig:meditation_remaining-6}
\end{minipage}
\end{paperfigure}
\clearpage

\begin{paperfigure}[p]
\centering
\sbox0{\includegraphics[width=0.485\textwidth,height=0.82\textheight,keepaspectratio]{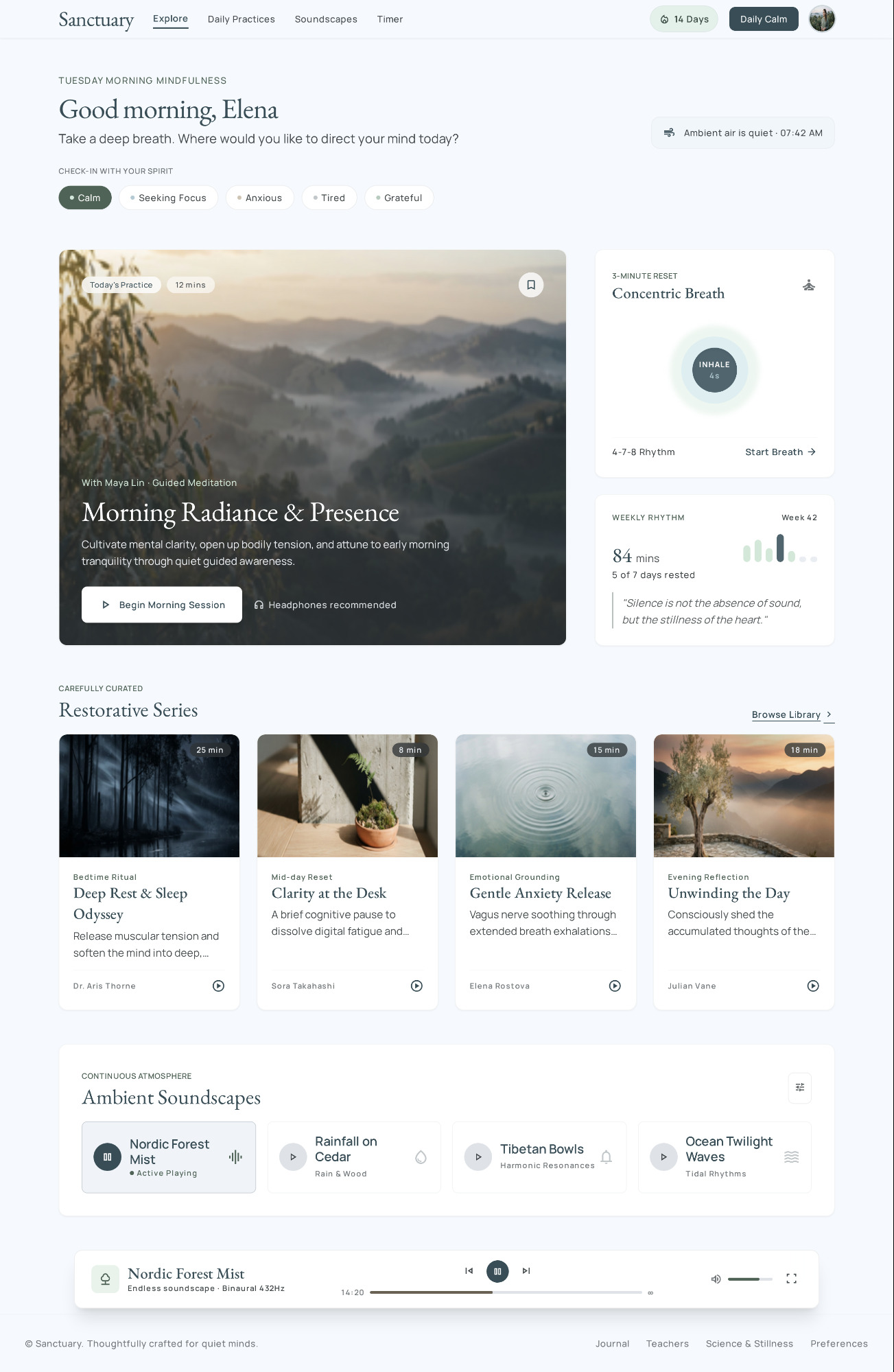}}
\sbox2{\includegraphics[width=0.485\textwidth,height=0.82\textheight,keepaspectratio]{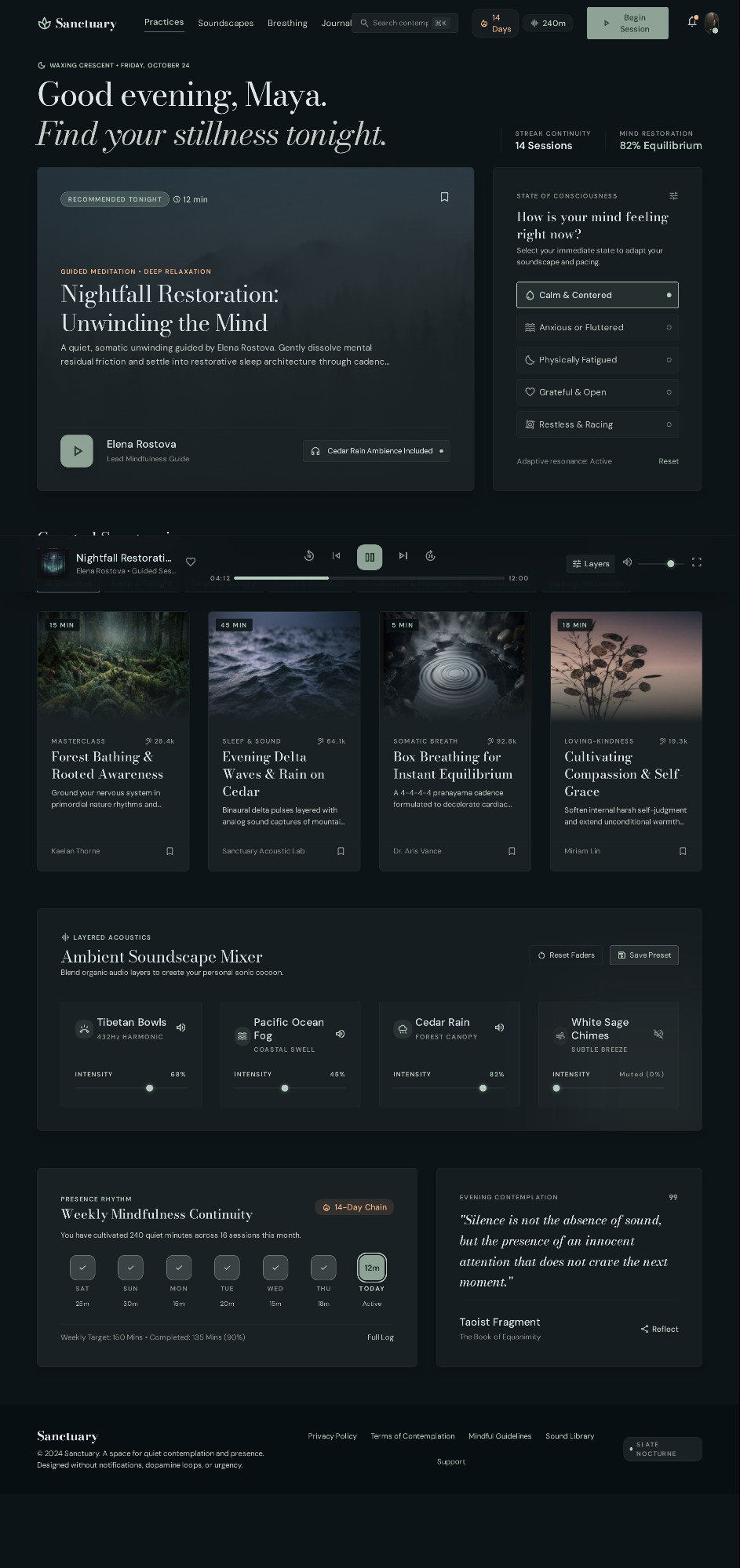}}
\dimen0=\ht0
\ifdim\ht2<\dimen0 \dimen0=\ht2\fi
\edef\appendixPairHeight{\the\dimen0}
\begin{minipage}[t]{0.485\textwidth}
\vspace{0pt}
\centering
\begin{minipage}[t][\appendixPairHeight][t]{\linewidth}
\centering
\includegraphics[height=\appendixPairHeight]{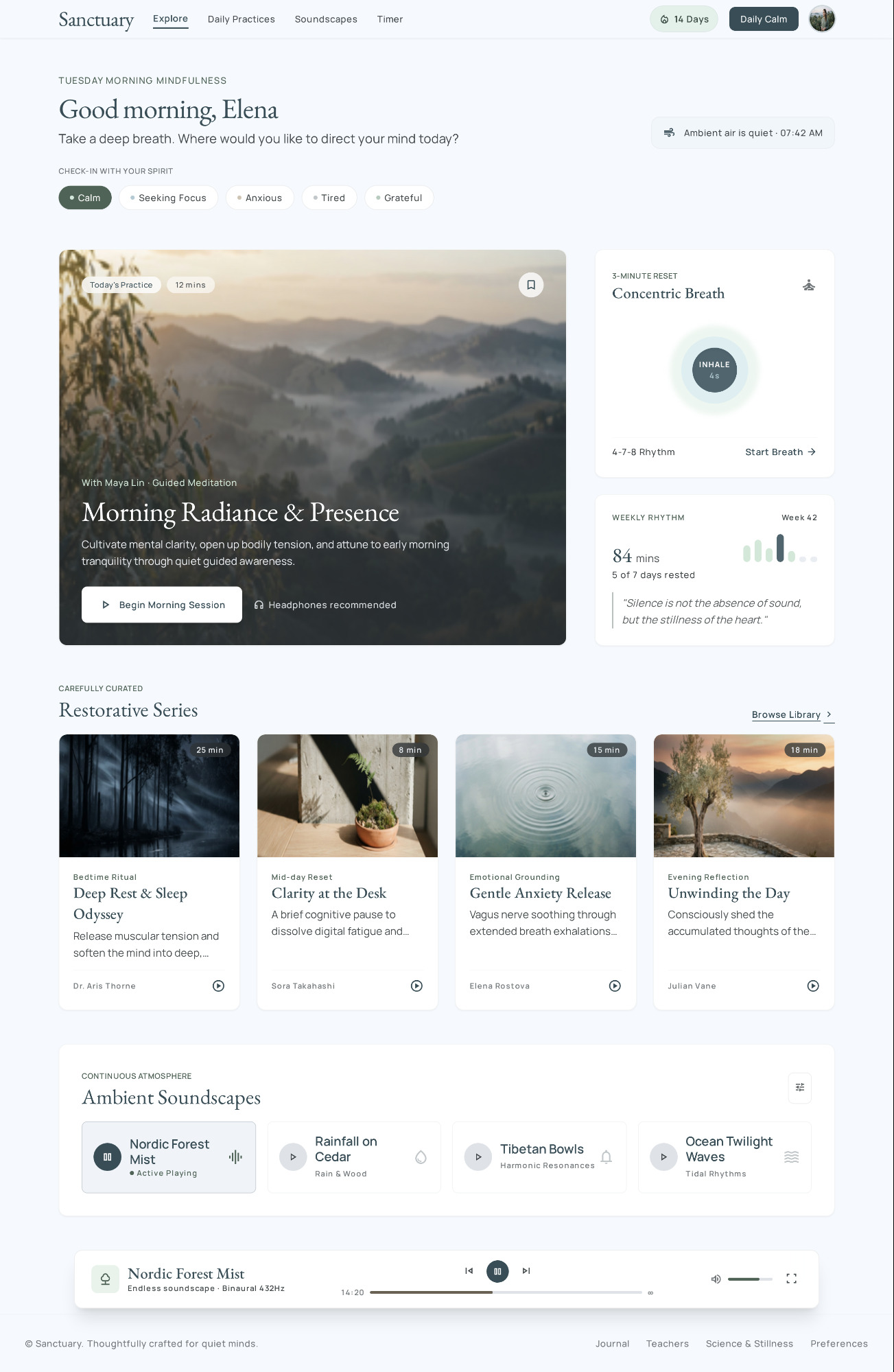}
\end{minipage}
\caption{Joint candidate sampling desktop meditation-app generation: Run 2 (\textit{Prem.}, $0.25$, light). Full screenshot.}
\label{fig:meditation_remaining-7}
\end{minipage}\hfill
\begin{minipage}[t]{0.485\textwidth}
\vspace{0pt}
\centering
\begin{minipage}[t][\appendixPairHeight][t]{\linewidth}
\centering
\includegraphics[height=\appendixPairHeight]{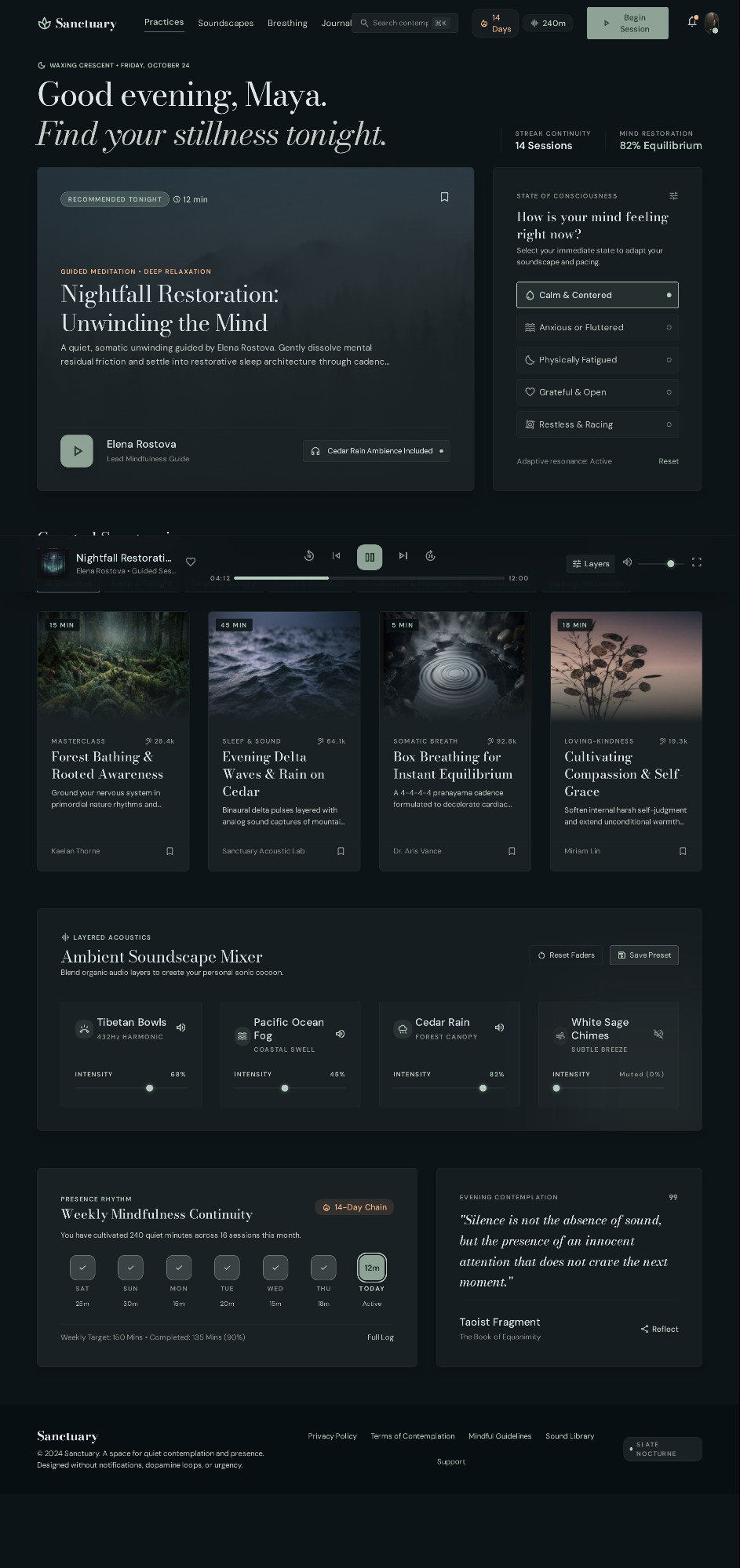}
\end{minipage}
\caption{Joint candidate sampling desktop meditation-app generation: Run 3 (\textit{Exp.}, $0.10$, \textbf{dark}). Full screenshot.}
\label{fig:meditation_remaining-8}
\end{minipage}
\end{paperfigure}
\clearpage

\begin{paperfigure}[p]
\centering
\sbox0{\includegraphics[width=0.485\textwidth,height=0.82\textheight,keepaspectratio]{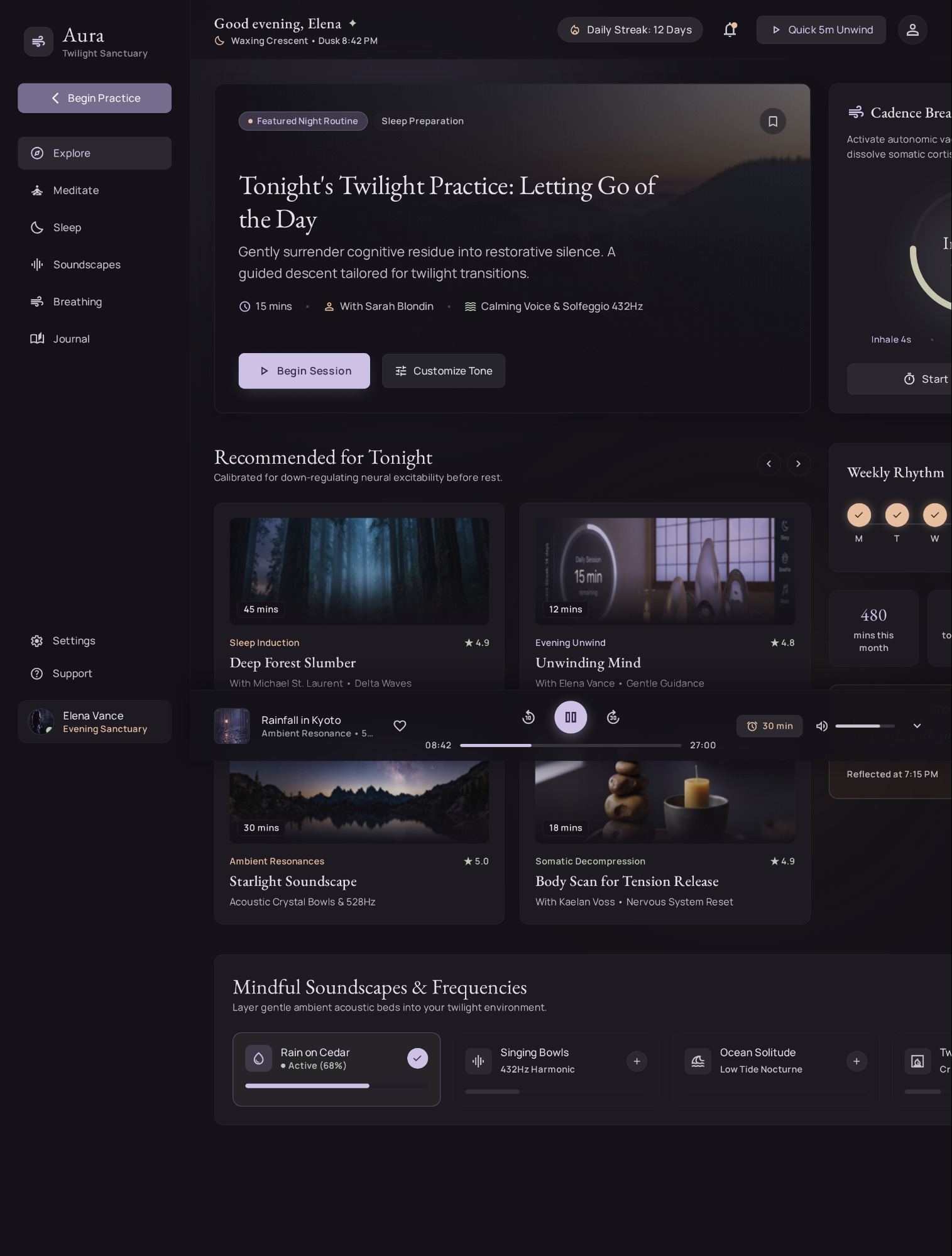}}
\sbox2{\includegraphics[width=0.485\textwidth,height=0.82\textheight,keepaspectratio]{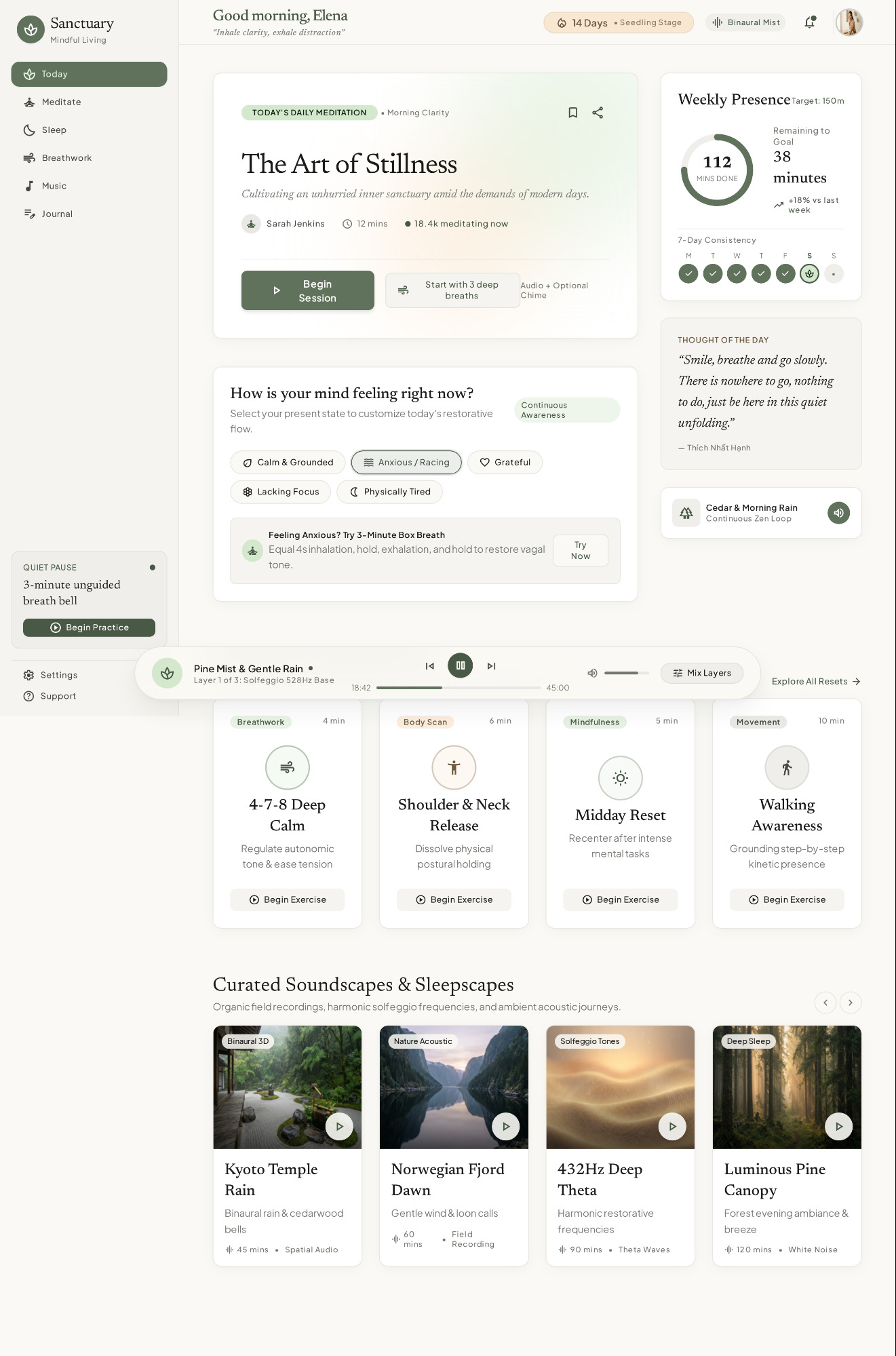}}
\dimen0=\ht0
\ifdim\ht2<\dimen0 \dimen0=\ht2\fi
\edef\appendixPairHeight{\the\dimen0}
\begin{minipage}[t]{0.485\textwidth}
\vspace{0pt}
\centering
\begin{minipage}[t][\appendixPairHeight][t]{\linewidth}
\centering
\includegraphics[height=\appendixPairHeight]{assets/optimized/meditation_vs/screen_4.jpg}
\end{minipage}
\caption{Joint candidate sampling desktop meditation-app generation: Run 4 (\textit{Prem.}, $0.25$, \textbf{dark}). Full screenshot.}
\label{fig:meditation_remaining-9}
\end{minipage}\hfill
\begin{minipage}[t]{0.485\textwidth}
\vspace{0pt}
\centering
\begin{minipage}[t][\appendixPairHeight][t]{\linewidth}
\centering
\includegraphics[height=\appendixPairHeight]{assets/optimized/meditation_vs/screen_5.jpg}
\end{minipage}
\caption{Joint candidate sampling desktop meditation-app generation: Run 5 (\textit{Safe}, $0.65$, light). Full screenshot.}
\label{fig:meditation_remaining-10}
\end{minipage}
\end{paperfigure}
\clearpage

\subsection{Mobile Meal-Planning Dashboard (5 vs.\ 5 Runs)}
\label{app:mealplan_case_study}

To check a denser layout on a mobile viewport, we ran five baseline and five joint-sampling generations for a meal-planning dashboard. Table~\ref{tab:token-diversity} compares design-token variety, Figures~\ref{fig:mealplan_remaining}--\ref{fig:mealplan_remaining-10} show all ten full screenshots in run order, and Table~\ref{tab:image-prompt-comparison} contrasts baseline image prompts against the three-tier sampled candidates.

\begin{papertable}[tp]
\centering
\caption{Design-token variety across five runs of the mobile meal-planning dashboard (percentages cover this single prompt only). Bold highlights shifts in color palette, headline typography, and corner radius.}
\label{tab:token-diversity}
\ifuseacl\small\else\footnotesize\fi
\setlength{\tabcolsep}{5pt}
\renewcommand{\arraystretch}{1.15}
\begin{tabular}{@{}lrr p{4.2cm}@{}}
\toprule
\textbf{Dimension} & \textbf{Baseline} & \textbf{Candidate sampling} & \textbf{Difference} \\
\midrule
\keycell{Primary Colors} & \keycell{2 (80\% \texttt{\#2D6A4F})} & \keycell{\textbf{5} (100\% unique)} & \keycell{Five distinct palettes} \\
\keycell{Headline Fonts} & \keycell{1 (100\% Plus Jakarta)} & \keycell{\textbf{3} unique} & \keycell{Adds Epilogue and Lexend} \\
Body Fonts & 2 (80\% Be Vietnam) & \textbf{3} unique & Adds Inter and Manrope pairings \\
\keycell{Corner Radius} & \keycell{1 (100\% level 2)} & \keycell{\textbf{3} unique} & \keycell{Spans sharp to rounded styles} \\
Lower-typicality picks & 0\% & \textbf{60\%} & 3 of 5 runs picked a $\hat p \le 0.40$ option \\
\bottomrule
\end{tabular}
\end{papertable}

\begin{papertable}[tp]
\centering
\caption{Qualitative comparison of generated image prompts for the mobile meal-planning-dashboard example. The selected direction is bold and marked with a dagger.}
\label{tab:image-prompt-comparison}
\scriptsize
\renewcommand{\arraystretch}{1.15}
\begin{tabularx}{\textwidth}{@{}l l p{0.18\textwidth} X@{}}
\toprule
\textbf{Method} & \textbf{Dish / target} & \textbf{Tier \& weight ($\hat p$)} & \textbf{Generated image prompt} \\
\midrule
\multirow{3}{*}{\textbf{Baseline}}
 & Grilled Salmon & Single prompt & \textit{``A beautifully plated grilled salmon fillet with crispy skin, resting on a bed of garlic-saut\'eed spinach and asparagus... atmospheric evening lighting, shallow depth of field, dark slate plate.''} \\
 & Avocado Toast & Single prompt & \textit{``A beautiful top-down shot of smashed avocado on rustic sourdough toast, topped with chili flakes, microgreens, and a poached egg breaking open. Bright morning sunlight, marble countertop.''} \\
 & Quinoa Salad & Single prompt & \textit{``Top-down view of a vibrant Mediterranean salad in a wide bowl, featuring sliced chicken, chickpeas, feta cheese crumbles, kalamata olives, cherry tomatoes... off-white background.''} \\
\midrule
\multirow{6}{*}{\textbf{\shortstack[l]{Candidate\\sampling}}}
 & \multirow{3}{*}{\shortstack[l]{Overnight Oats\\(Breakfast)}}
   & \textit{Safe} ($\hat p=0.60$) & \textit{``Close-up photograph of a clear glass mason jar filled with overnight oats, topped with fresh blueberries, sliced bananas, chia seeds... soft morning light from the side.''} \\
& & \keycell{\textbf{\textit{Premium}} ($\hat p=0.25$)\textsuperscript{\textdagger}} & \keycell{\textit{``Elevated, editorial-style photograph of layered overnight oats on a light oak wood surface, distinctly visible layers through clear glass, dappled sunlight, shallow depth of field.''}} \\
 & & \textit{Experimental} ($\hat p=0.15$) & \textit{``Artistic, high-key photograph of a deconstructed overnight oats experience in a minimalist, handmade ceramic bowl, surrounded by artfully scattered ingredients on raw linen.''} \\
\cmidrule{2-4}
 & \multirow{3}{*}{\shortstack[l]{Lentil Stew\\(Dinner)}}
   & \textit{Safe} ($\hat p=0.60$) & \textit{``Close-up photograph of a steaming, rustic ceramic bowl filled with hearty lentil and root vegetable stew on a worn wooden dining table, garnished with fresh parsley.''} \\
 & & \textbf{\textit{Premium}} ($\hat p=0.25$) & \textit{``Elevated, moody culinary photograph of artisanal stoneware containing a rich, slow-cooked lentil and root vegetable ragout under soft directional side-lighting and olive oil drizzle.''} \\
 & & \textit{Experimental} ($\hat p=0.15$) & \textit{``Top-down, hyper-realistic still life composition set in a minimalist, Scandinavian-inspired kitchen on light bleached wood with a dark glazed bowl of steaming stew.''} \\
\bottomrule
\end{tabularx}
\vspace{1mm}
{\raggedright \textsuperscript{\textdagger} Direction sampled under $\tau=2.0$ external selection for this run.\par}
\end{papertable}

Across both prompts, joint sampling varies palette, light/dark mode, typography, corner radius, and imagery together. These two five-run comparisons illustrate the visual shift concretely alongside the quantitative benchmarks in Section~\ref{sec:results}.

\clearpage
\begin{paperfigure}[p]
\centering
\sbox0{\includegraphics[width=0.485\textwidth,height=0.82\textheight,keepaspectratio]{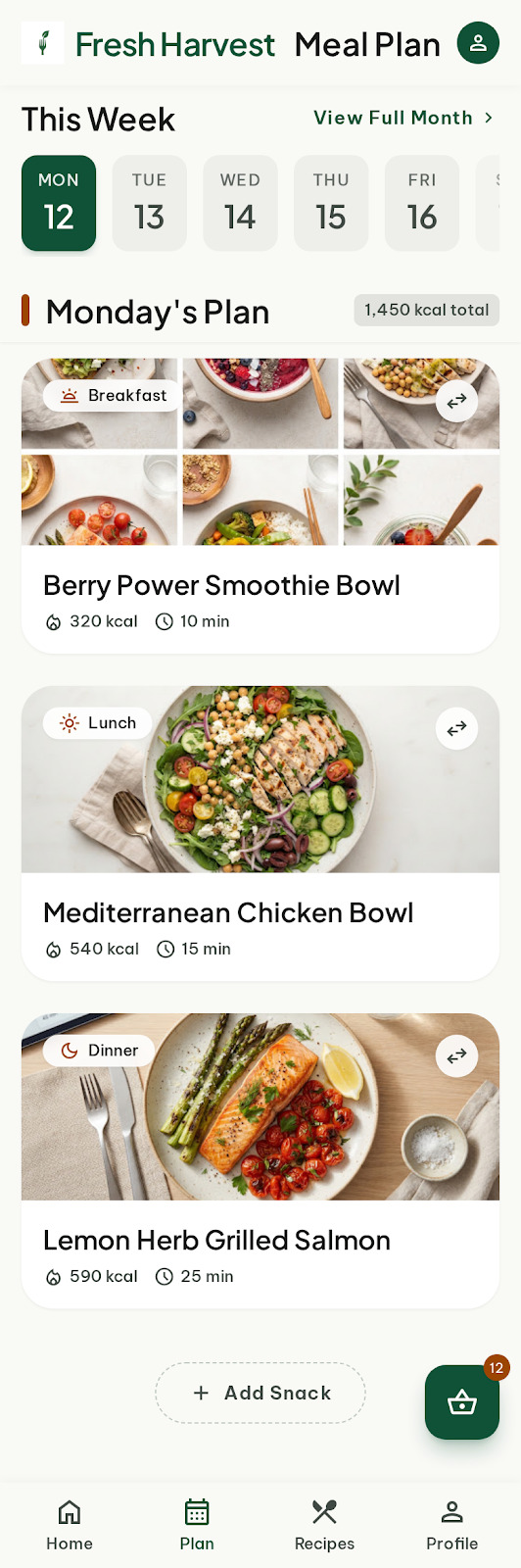}}
\sbox2{\includegraphics[width=0.485\textwidth,height=0.82\textheight,keepaspectratio]{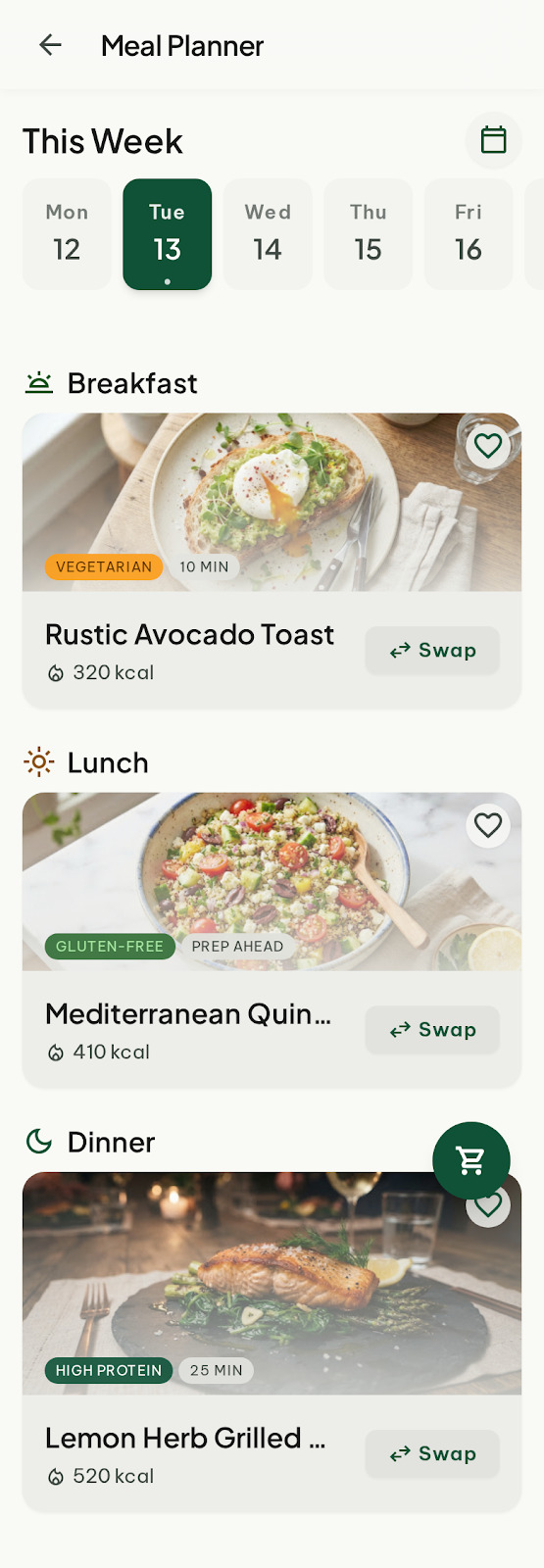}}
\dimen0=\ht0
\ifdim\ht2<\dimen0 \dimen0=\ht2\fi
\edef\appendixPairHeight{\the\dimen0}
\begin{minipage}[t]{0.485\textwidth}
\vspace{0pt}
\centering
\begin{minipage}[t][\appendixPairHeight][t]{\linewidth}
\centering
\includegraphics[height=\appendixPairHeight]{assets/optimized/mealplan_no_vs/screen_1.jpg}
\end{minipage}
\caption{Baseline mobile meal-planning-dashboard generation: Run 1. Full screenshot.}
\label{fig:mealplan_remaining}
\end{minipage}\hfill
\begin{minipage}[t]{0.485\textwidth}
\vspace{0pt}
\centering
\begin{minipage}[t][\appendixPairHeight][t]{\linewidth}
\centering
\includegraphics[height=\appendixPairHeight]{assets/optimized/mealplan_no_vs/screen_3.jpg}
\end{minipage}
\caption{Baseline mobile meal-planning-dashboard generation: Run 2. Full screenshot.}
\label{fig:mealplan_remaining-2}
\end{minipage}
\end{paperfigure}
\clearpage

\begin{paperfigure}[p]
\centering
\sbox0{\includegraphics[width=0.485\textwidth,height=0.82\textheight,keepaspectratio]{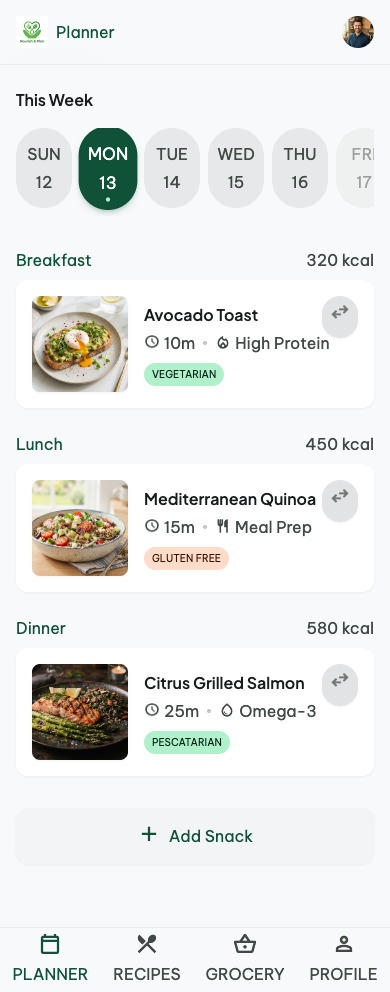}}
\sbox2{\includegraphics[width=0.485\textwidth,height=0.82\textheight,keepaspectratio]{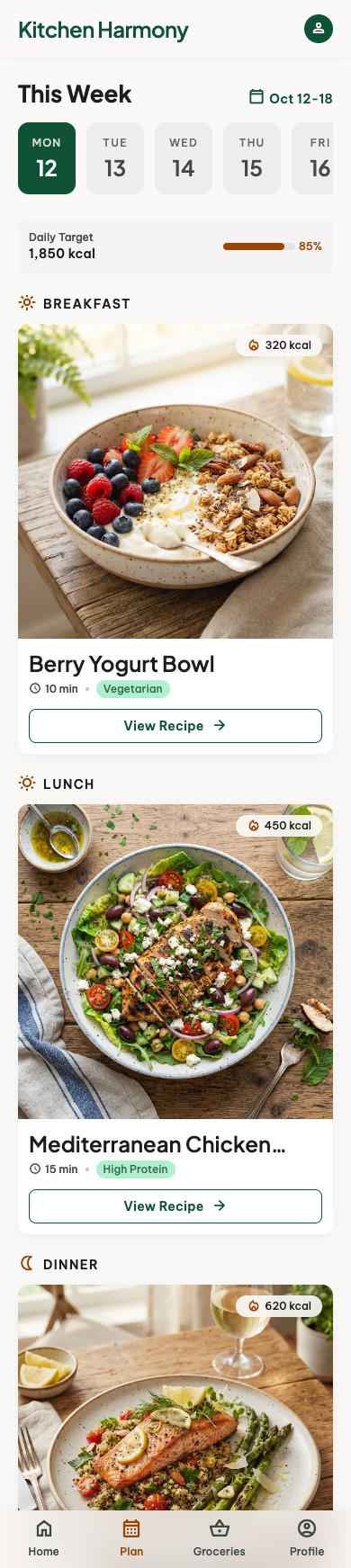}}
\dimen0=\ht0
\ifdim\ht2<\dimen0 \dimen0=\ht2\fi
\edef\appendixPairHeight{\the\dimen0}
\begin{minipage}[t]{0.485\textwidth}
\vspace{0pt}
\centering
\begin{minipage}[t][\appendixPairHeight][t]{\linewidth}
\centering
\includegraphics[height=\appendixPairHeight]{assets/optimized/mealplan_no_vs/screen_2.jpg}
\end{minipage}
\caption{Baseline mobile meal-planning-dashboard generation: Run 3. Full screenshot.}
\label{fig:mealplan_remaining-3}
\end{minipage}\hfill
\begin{minipage}[t]{0.485\textwidth}
\vspace{0pt}
\centering
\begin{minipage}[t][\appendixPairHeight][t]{\linewidth}
\centering
\includegraphics[height=\appendixPairHeight]{assets/optimized/mealplan_no_vs/screen_4.jpg}
\end{minipage}
\caption{Baseline mobile meal-planning-dashboard generation: Run 4. Full screenshot.}
\label{fig:mealplan_remaining-4}
\end{minipage}
\end{paperfigure}
\clearpage

\begin{paperfigure}[p]
\centering
\sbox0{\includegraphics[width=0.485\textwidth,height=0.82\textheight,keepaspectratio]{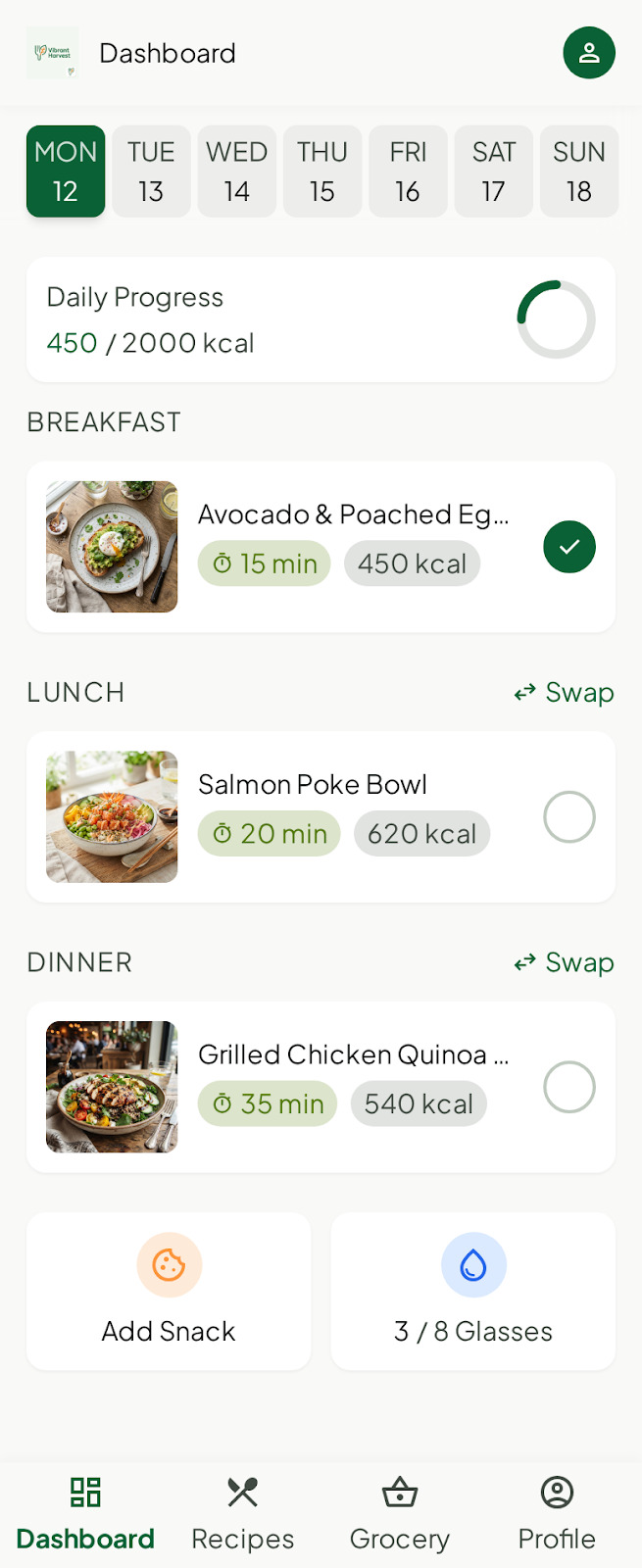}}
\sbox2{\includegraphics[width=0.485\textwidth,height=0.82\textheight,keepaspectratio]{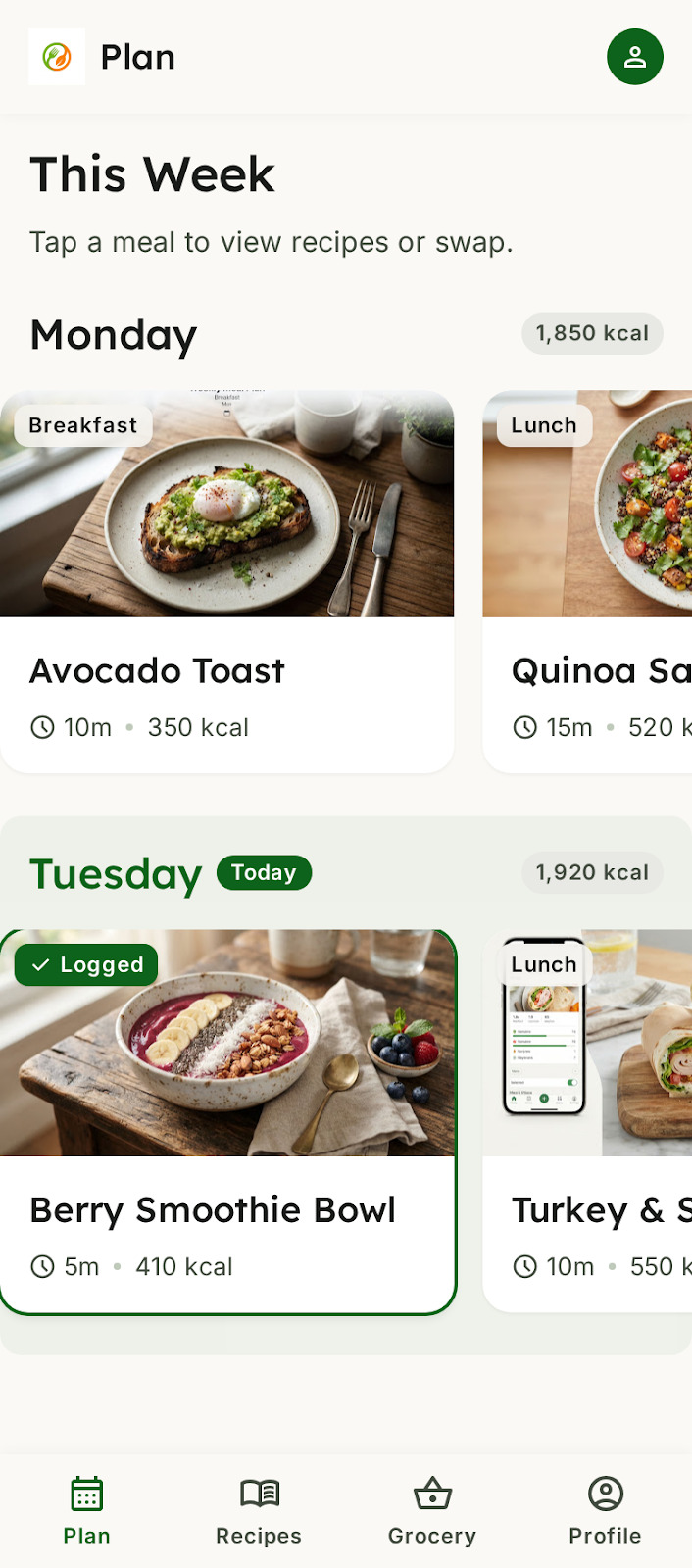}}
\dimen0=\ht0
\ifdim\ht2<\dimen0 \dimen0=\ht2\fi
\edef\appendixPairHeight{\the\dimen0}
\begin{minipage}[t]{0.485\textwidth}
\vspace{0pt}
\centering
\begin{minipage}[t][\appendixPairHeight][t]{\linewidth}
\centering
\includegraphics[height=\appendixPairHeight]{assets/optimized/mealplan_no_vs/screen_5.jpg}
\end{minipage}
\caption{Baseline mobile meal-planning-dashboard generation: Run 5. Full screenshot.}
\label{fig:mealplan_remaining-5}
\end{minipage}\hfill
\begin{minipage}[t]{0.485\textwidth}
\vspace{0pt}
\centering
\begin{minipage}[t][\appendixPairHeight][t]{\linewidth}
\centering
\includegraphics[height=\appendixPairHeight]{assets/optimized/mealplan_vs/screen_3.jpg}
\end{minipage}
\caption{Joint candidate sampling mobile meal-planning-dashboard generation: Run 1. Full screenshot.}
\label{fig:mealplan_remaining-6}
\end{minipage}
\end{paperfigure}
\clearpage

\begin{paperfigure}[p]
\centering
\sbox0{\includegraphics[width=0.485\textwidth,height=0.82\textheight,keepaspectratio]{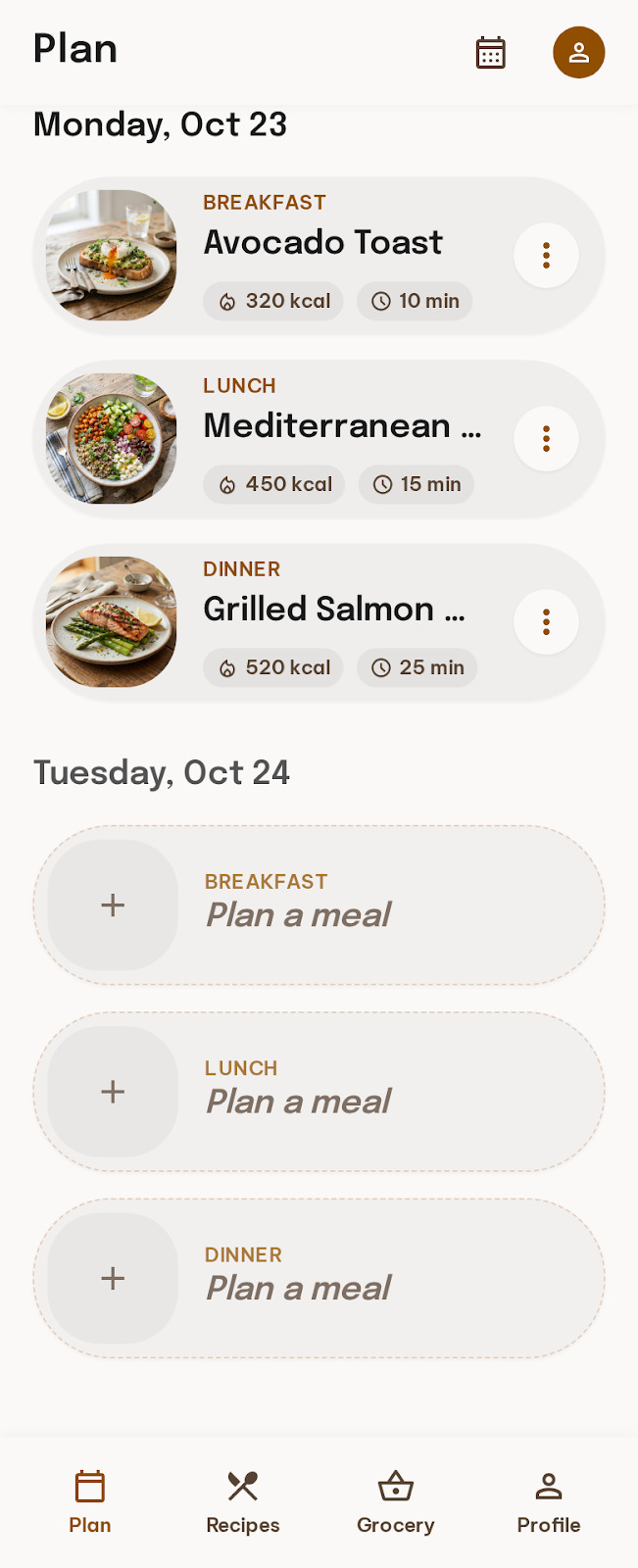}}
\sbox2{\includegraphics[width=0.485\textwidth,height=0.82\textheight,keepaspectratio]{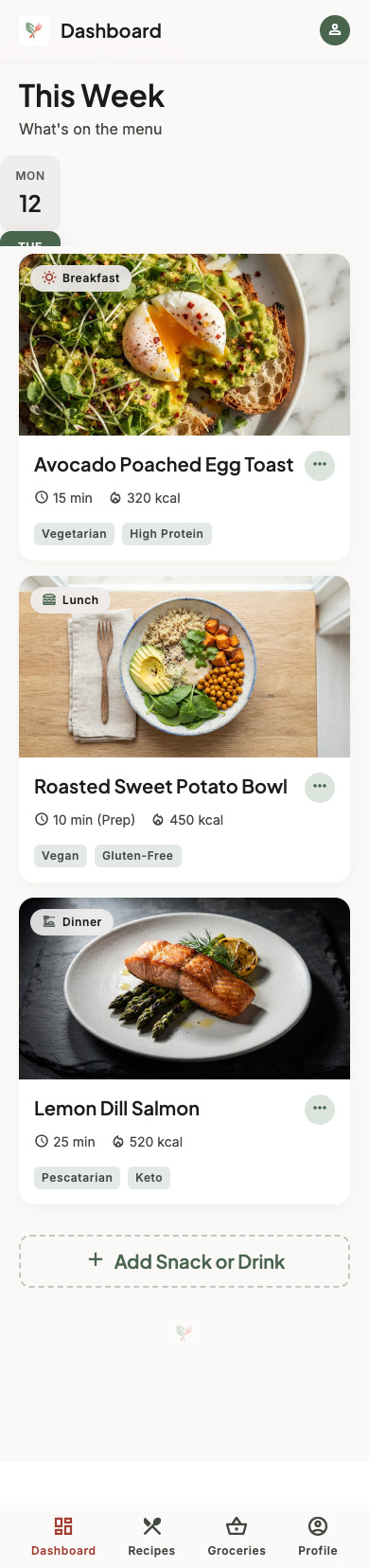}}
\dimen0=\ht0
\ifdim\ht2<\dimen0 \dimen0=\ht2\fi
\edef\appendixPairHeight{\the\dimen0}
\begin{minipage}[t]{0.485\textwidth}
\vspace{0pt}
\centering
\begin{minipage}[t][\appendixPairHeight][t]{\linewidth}
\centering
\includegraphics[height=\appendixPairHeight]{assets/optimized/mealplan_vs/screen_5.jpg}
\end{minipage}
\caption{Joint candidate sampling mobile meal-planning-dashboard generation: Run 2. Full screenshot.}
\label{fig:mealplan_remaining-7}
\end{minipage}\hfill
\begin{minipage}[t]{0.485\textwidth}
\vspace{0pt}
\centering
\begin{minipage}[t][\appendixPairHeight][t]{\linewidth}
\centering
\includegraphics[height=\appendixPairHeight]{assets/optimized/mealplan_vs/screen_1.jpg}
\end{minipage}
\caption{Joint candidate sampling mobile meal-planning-dashboard generation: Run 3. Full screenshot.}
\label{fig:mealplan_remaining-8}
\end{minipage}
\end{paperfigure}
\clearpage

\begin{paperfigure}[p]
\centering
\sbox0{\includegraphics[width=0.485\textwidth,height=0.82\textheight,keepaspectratio]{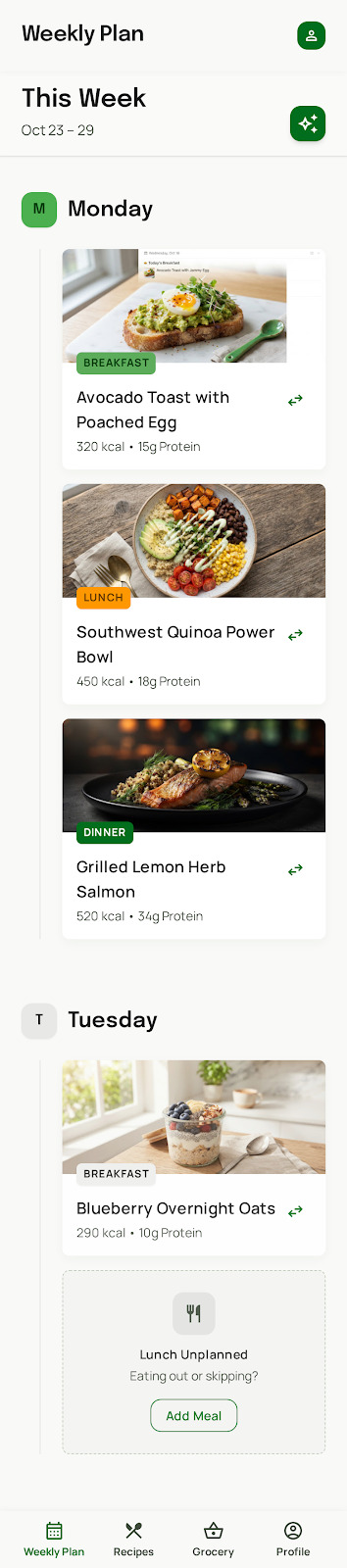}}
\sbox2{\includegraphics[width=0.485\textwidth,height=0.82\textheight,keepaspectratio]{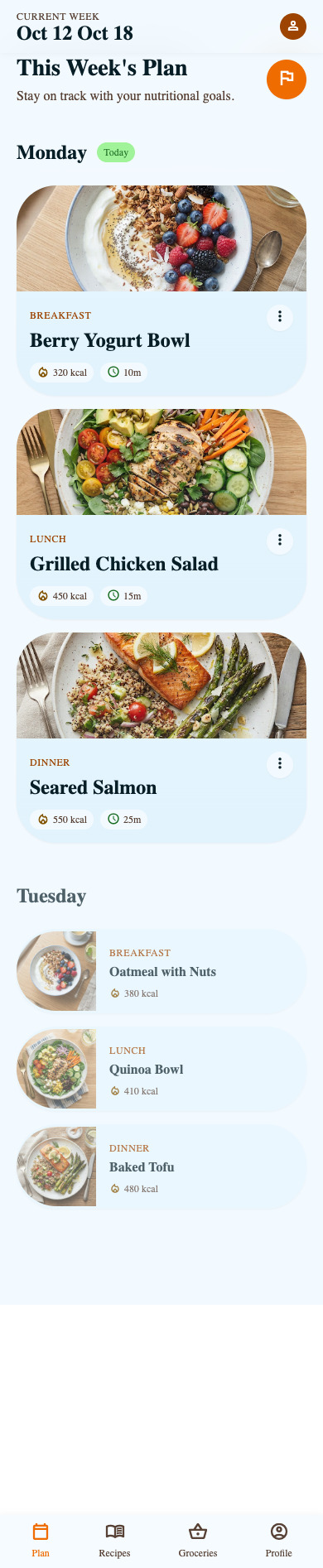}}
\dimen0=\ht0
\ifdim\ht2<\dimen0 \dimen0=\ht2\fi
\edef\appendixPairHeight{\the\dimen0}
\begin{minipage}[t]{0.485\textwidth}
\vspace{0pt}
\centering
\begin{minipage}[t][\appendixPairHeight][t]{\linewidth}
\centering
\includegraphics[height=\appendixPairHeight]{assets/optimized/mealplan_vs/screen_2.jpg}
\end{minipage}
\caption{Joint candidate sampling mobile meal-planning-dashboard generation: Run 4. Full screenshot.}
\label{fig:mealplan_remaining-9}
\end{minipage}\hfill
\begin{minipage}[t]{0.485\textwidth}
\vspace{0pt}
\centering
\begin{minipage}[t][\appendixPairHeight][t]{\linewidth}
\centering
\includegraphics[height=\appendixPairHeight]{assets/optimized/mealplan_vs/screen_4.jpg}
\end{minipage}
\caption{Joint candidate sampling mobile meal-planning-dashboard generation: Run 5. Full screenshot.}
\label{fig:mealplan_remaining-10}
\end{minipage}
\end{paperfigure}
\clearpage

\ifuseacl
  \clearpage
  \endgroup
  \twocolumn
  \input{sections/appendix_reproducibility.tex}
\fi

\ifuseneurips
  \clearpage
  \input{checklist.tex}
\fi

\end{document}